\pdfoutput=1
\documentclass{article}

\usepackage[accepted]{icml2024}

\usepackage{soul}

\usepackage{pdfpages}
\usepackage{amsmath}
\usepackage{amsfonts}
\usepackage{mathtools}
\usepackage{amsthm}

\usepackage{microtype}
\usepackage{graphicx}
\usepackage{booktabs} 
\usepackage{algorithm}
\usepackage{algorithmic}
\usepackage[backref=page,colorlinks=true,linkcolor=blue,citecolor=green,urlcolor=blue]{hyperref}
\usepackage{url}
\usepackage{graphicx}
\usepackage{bbm}
\usepackage{placeins}
\usepackage{adjustbox}
\usepackage{xcolor}
\usepackage{scalerel}
\usepackage{natbib}
\usepackage[outdir=./]{epstopdf}
\usepackage{graphicx,float,pgfplots,wrapfig,sidecap,lipsum}

\usepackage{colortbl}
\usepackage{diagbox} 
\usepackage{tablefootnote}
\usepackage[font=small,labelfont=bf]{caption}
\usepackage{subcaption}
\usepackage{subfloat}
\usepackage{tikz}
\usetikzlibrary{fit}
\usetikzlibrary{calc,shapes}
\usetikzlibrary{decorations.pathmorphing} 
\usetikzlibrary{fit}					
\usetikzlibrary{backgrounds}	
\usetikzlibrary{pgfplots.groupplots}

\usepackage[utf8]{inputenc}
\usepackage{pgfplots}
\usepgfplotslibrary{groupplots,dateplot}
\usetikzlibrary{patterns,shapes.arrows}
\pgfplotsset{compat=newest}

\usepackage{xspace}
\usepackage{soul}
\usepackage{booktabs}
\usepackage{bm}
	
\usepackage[capitalise]{cleveref}

\usepackage{listings}
\usepackage{cleveref}
\usepackage{multirow}
\usepackage{xcolor}

\usepackage{enumitem}

\definecolor{codegreen}{rgb}{0,0.6,0}
\definecolor{codegray}{rgb}{0.5,0.5,0.5}
\definecolor{codepurple}{rgb}{0.58,0,0.82}
\definecolor{backcolour}{rgb}{0.95,0.95,0.92}

\lstdefinestyle{mystyle}{
    backgroundcolor=\color{backcolour},   
    commentstyle=\color{codegreen},
    keywordstyle=\color{magenta},
    numberstyle=\tiny\color{codegray},
    stringstyle=\color{codepurple},
    basicstyle=\ttfamily\footnotesize,
    breakatwhitespace=false,         
    breaklines=true,                 
    captionpos=b,                    
    keepspaces=true,                 
    numbers=left,                    
    numbersep=5pt,                  
    showspaces=false,                
    showstringspaces=false,
    showtabs=false,                  
    tabsize=2
}

\theoremstyle{remark}

\AtBeginEnvironment{proof}{}{}{}

\newcommand{\ours}{Premier-TACO\xspace}
\newcommand{\mw}{MetaWorld\xspace}
\newcommand{\dmc}{Deepmind Control Suite\xspace}
\newcommand{\lfs}{Learn-from-scratch\xspace}

\definecolor{sourcecolor}{rgb}{0.5,1,0.5}
\definecolor{ourcolor}{rgb}{1,0,0}
\definecolor{singlecolor}{rgb}{0,0,1}
\definecolor{auxcolor}{rgb}{0.54,0.17,0.89}
\definecolor{linearcolor}{rgb}{0.172549019607843,0.627450980392157,0.172549019607843}
\definecolor{randomcolor}{rgb}{1,0.498039215686275,0.0549019607843137}
\definecolor{tunecolor}{rgb}{0.9568627450980393, 0.8156862745098039,0}
\definecolor{aligncolor}{rgb}{0,0.5,0}

\usepackage[disable,textsize=tiny]{todonotes}
\usepackage[textsize=tiny]{todonotes}

\icmltitlerunning{Premier-TACO: Pretraining Multitask Representation via Temporal Action-Driven Contrastive Loss}

\begin{document}

\twocolumn[
\icmltitle{Premier-TACO is a Few-Shot Policy Learner: Pretraining Multitask Representation via Temporal Action-Driven Contrastive Loss}

\icmlsetsymbol{equal}{*}

\begin{icmlauthorlist}
\icmlauthor{Ruijie Zheng}{umd}
\icmlauthor{Yongyuan Liang}{umd}
\icmlauthor{Xiyao Wang}{umd}
\icmlauthor{Shuang Ma}{microsoft}\\
\icmlauthor{Hal Daumé III}{umd}
\icmlauthor{Huazhe Xu}{tsu}
\icmlauthor{John Langford}{microsoft}\\
\icmlauthor{Praveen Palanisamy}{microsoft}
\icmlauthor{Kalyan Shankar Basu}{microsoft}
\icmlauthor{Furong Huang}{umd}
\end{icmlauthorlist}

\icmlaffiliation{umd}{Department of Computer Science, University of Maryland, College Park}
\icmlaffiliation{tsu}{Tsinghua University}
\icmlaffiliation{microsoft}{Microsoft Research}

\icmlcorrespondingauthor{Ruijie Zheng}{rzheng12@umd.edu}
\icmlkeywords{sequential decision making, RL, multitask offline pretraining}
\vskip 0.3in
]



\printAffiliationsAndNotice{}  
\begin{abstract}
We present Premier-TACO, a multitask feature representation learning approach designed to improve few-shot policy learning efficiency in sequential decision-making tasks. Premier-TACO leverages a subset of multitask offline datasets for pretraining a \textit{general feature representation}, which captures critical environmental dynamics and is fine-tuned using minimal expert demonstrations.  It advances the temporal action contrastive learning (TACO) objective, known for state-of-the-art results in visual control tasks, by incorporating a novel negative example sampling strategy. This strategy is crucial in significantly boosting TACO's computational efficiency, making large-scale multitask offline pretraining feasible. 
Our extensive empirical evaluation in a diverse set of continuous control benchmarks including Deepmind Control Suite, MetaWorld, and LIBERO demonstrate Premier-TACO's effectiveness in pretraining visual representations, significantly enhancing few-shot imitation learning of novel tasks. 
Our code, pretraining data, as well as pretrained model checkpoints will be released at \href{https://github.com/PremierTACO/premier-taco}{https://github.com/PremierTACO/premier-taco}.
\end{abstract}

\section{Introduction}

In the dynamic and ever-changing world we inhabit, the importance of sequential decision-making (SDM) in machine learning cannot be overstated. 
Unlike static tasks, sequential decisions reflect the fluidity of real-world scenarios, from robotic manipulations to evolving healthcare treatments. 
Just as foundation models in language, such as BERT~\citep{bert} and GPT~\citep{gpt2, gpt3}, have revolutionized natural language processing by leveraging vast amounts of textual data to understand linguistic nuances, \textit{pretrained foundation models} hold similar promise for sequential decision-making (SDM). 
In language, these models capture the essence of syntax, semantics, and context, serving as a robust starting point for a myriad of downstream tasks. Analogously, in SDM, where decisions are influenced by a complex interplay 
\begin{figure}[t!]
\centering
\includegraphics[width=1.0\columnwidth]{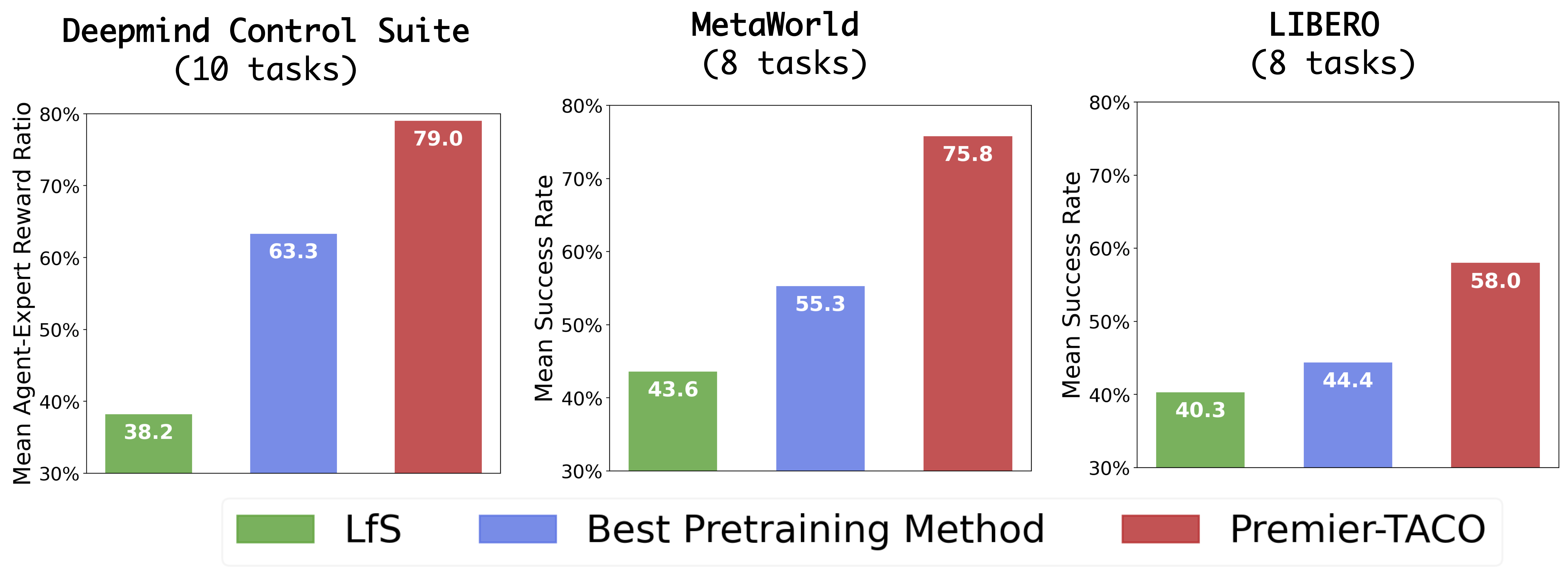}
 \caption{Performance of \ours pretrained visual representation for few-shot imitation learning on downstream unseen tasks from Deepmind Control Suite, MetaWorld, and LIBERO. LfS here represents \textit{learning from scratch.}}\label{fig:averaged-main-result}
 \end{figure}
 of past actions, current states, and future possibilities, a pretrained foundation model can provide a rich, generalized understanding of decision sequences. This foundational knowledge, built upon diverse decision-making scenarios, can then be fine-tuned to specific tasks, much like how language models are adapted to specific linguistic tasks.

The following \textbf{challenges} are unique to sequential decision-making, setting it apart from existing vision and language pretraining paradigms.
\textbf{(C1) Data Distribution Shift}: Training data usually consists of specific behavior-policy-generated trajectories. This leads to vastly different data distributions at various stages—pretraining, finetuning, and deployment—resulting in compromised performance~\citep{lee2021offlinetoonline}. \textbf{(C2) Task Heterogeneity}: Unlike language and vision tasks, which often share semantic features, decision-making tasks vary widely in configurations, transition dynamics, and state and action spaces. This makes it difficult to develop a universally applicable representation. \textbf{(C3) Data Quality and Supervision}: Effective representation learning often relies on high-quality data and expert guidance. However, these resources are either absent or too costly to obtain in many real-world decision-making tasks~\citep{brohan2023rt1,stooke2021decoupling}.
Our \textbf{aspirational criteria} for foundation model for sequential decision-making encompass several key features: \textbf{(W1) Versatility} that allows the model to generalize across a wide array of tasks, even those not previously encountered, such as new embodiments viewed or observations from novel camera angles; \textbf{(W2) Efficiency} in adapting to downstream tasks, requiring minimal data through few-shot learning techniques; \textbf{(W3) Robustness} to pretraining data of fluctuating quality, ensuring a resilient foundation; and 
\textbf{(W4) Compatibility} with existing large pretrained models such as ~\citet{nair2022r3m}.  \looseness=-1

In light of these challenges and desirables in building foundation models for SDM, our approach to develop foundational models for sequential decision-making focuses on creating a universal and transferable encoder using a reward-free, dynamics based, temporal contrastive pretraining objective. 
This encoder would be tailored to manage tasks with complex observation spaces, such as visual inputs. 
By excluding reward signals during the pretraining stage, the model will be better poised to generalize across a broad array of downstream tasks that may have divergent objectives. 
Leveraging a world-model approach ensures that the encoder learns a compact representation that can capture universal transition dynamics, akin to the laws of physics, thereby making it adaptable for multiple scenarios.
This encoder enables the transfer of knowledge to downstream control tasks, even when such tasks were not part of the original pretraining data set.





Existing works apply self-supervised pre-training from rich vision data such as ImageNet~\citep{deng2009imagenet} or Ego4D datasets~\citep{grauman2022ego4d} to build foundation models~\citep{nair2022r3m,majumdar2023vc1, ma2023vip}.
However, applying these approaches to sequential decision-making tasks is challenging. 
Specifically, they often overlook control-relevant considerations and suffer from a domain gap between pre-training datasets and downstream control tasks. 
In this paper, rather than focusing on leveraging large vision datasets, we propose a novel control-centric objective function for pretraining. 
Our approach, called \textit{\ours} (\textit{\underline{pre}training \underline{m}ult\underline{i}task r\underline{e}p\underline{r}esentation via \underline{t}emporal \underline{a}ction-driven \underline{co}ntrastive loss}), employs a temporal action-driven contrastive loss function for pretraining. 
This control-centric objective learns a state representation by optimizing the mutual information between representations of current states paired with action sequences and representations of the corresponding future states.\looseness=-1

\ours markedly enhances the effectiveness and efficiency of the temporal action contrastive learning (TACO) objective, as detailed in~\citet{zheng2023taco}, which delivers state-of-the-art outcomes in visual control tasks within a \textit{single-task setting}. It extends these capabilities to efficient, large-scale \textit{multitask offline pretraining}, broadening its applicability and performance.
Specifically, while TACO considers every data point in a batch as a potential negative example, \ours strategically samples a single negative example from a proximate window of the subsequent state. This method ensures the negative example is visually akin to the positive one, necessitating that the latent representation captures control-relevant information, rather than relying on extraneous features like visual appearance. This efficient negative example sampling strategy adds no computational burden and is compatible with smaller batch sizes. 
In particular, on \mw, using a batch size of $\frac{1}{8}$ for TACO, \ours achieves a 25\% relative performance improvement.
\ours can be seamlessly scaled for multitask offline pretraining, enhancing its usability and effectiveness.

Below we list our key contributions:
\begin{list}{$\rhd$}{\topsep=0.ex \leftmargin=0.2in \rightmargin=0.in \itemsep =0.0in}
\item  \textbf{(1)} We introduce \ours, a new framework designed for the multi-task offline visual representation pretraining of sequential decision-making problems. In particular, we develop a new temporal contrastive learning objective within the \ours framework. Compared with other temporal contrastive learning objectives such as TACO, \ours employs a simple yet efficient negative example sampling strategy, making it computationally feasible for multi-task representation learning.
\item  \textbf{(2) [(W1) Versatility (W2) Efficiency]} Through extensive empirical evaluation, we verify the effectiveness of \ours's pretrained visual representations for few-shot learning on unseen tasks. 
On \mw~\citep{metaworld} and LIBERO~\cite{libero}, with 5 expert trajectories, \ours outperforms the best baseline pretraining method by 37\% and 17\% respectively. 
Remarkably, in LIBERO, we are the first method to demonstrate benefits from pretraining.
On Deepmind Control Suite (DMC)~\citep{dmc}, using only 20 trajectories, which is considerably fewer demonstrations than~\citep{sun2023smart, majumdar2023vc1}, \ours achieves the best performance across 10 challenging tasks, including the hard Dog and Humanoid tasks.
This versatility extends even to unseen embodiments in DMC as well as unseen tasks with unseen camera views in \mw.

\item  \textbf{(3) [(W3) Robustness (W4) Compatability]} 
Furthermore, we demonstrate that \ours is not only resilient to data of lower quality but also compatible with exisiting large pretrained models. In DMC, \ours works well with the pretraining dataset collected randomly. 
Additionally, we showcase the capability of the temporal contrastive learning objective of \ours to finetune a generalized visual encoder such as R3M~\citep{nair2022r3m}, resulting in an averaged performance enhancement of around 50\% across the assessed tasks on Deepmind Control Suite and MetaWorld.

\end{list}


\section{Preliminary}


\subsection{Multitask Offline Pretraining}
We consider a collection of tasks $\big\{\mathcal{T}_i:(\mathcal{X}, \mathcal{A}_i, \mathcal{P}_i, \mathcal{R}_i, \gamma)\big\}_{i=1}^{N}$ with the same dimensionality in observation space $\mathcal{X}$. Let $\phi: \mathcal{X}\rightarrow \mathcal{Z}$ be a representation function of the agent's observation, which is either randomly initialized or pre-trained already on a large-scale vision dataset such as ImageNet~\citep{deng2009imagenet} or Ego4D~\citep{grauman2022ego4d}. 
Assuming that the agent is given a multitask offline dataset $\{(x_i, a_i, x'_i, r_i)\}$ of a subset of $K$ tasks $\{\mathcal{T}_{n_j}\}_{j=1}^{K}$. 
The objective is to pretrain a generalizable state representation $\phi$ or a motor policy $\pi$ so that when facing an unseen downstream task, it could quickly adapt with few expert demonstrations, using the pretrained representation. \newline
Below we summarize the pretraining and finetuning setups.\newline
\textbf{Pretraining}: The agent get access to a multitask offline dataset, which could be highly suboptimal. The goal is to learn a generalizable shared state representation from pixel inputs.\newline
\textbf{Adaptation}: Adapt to unseen downstream task from few expert demonstration with imitation learning. \looseness=-1

\subsection{TACO: Temporal Action Driven Contrastive Learning Objective}

Temporal Action-driven Contrastive Learning (TACO)~\citep{zheng2023taco} is a reinforcement learning algorithm proposed for addressing the representation learning problem in visual continuous control.
It aims to maximize the mutual information between representations of current states paired with action sequences and representations of the corresponding future states:
\begin{equation}
\label{eq:mi_obj}
\mathbb{J}_{\text{TACO}}=\mathcal{I}(Z_{t+K}; [Z_t,U_t,..., U_{t+K-1}])
\end{equation}
Here, $Z_t=\phi(X_t)$ and $U_t=\psi(A_t)$ represent latent state and action variables. 
Theoretically, it could be shown that maximization of this mutual information objective lead to state and action representations that are capable of representing the optimal value functions.
Empirically, TACO estimate the lower bound of the mutual information objective by the InfoNCE loss, and it achieves the state of art performance for both online and offline visual continuous control, demonstrating the effectiveness of temporal contrastive learning for representation learning in sequential decision making problems. \looseness=-1

\section{Method}

We introduce \ours, a generalized pre-training approach specifically formulated to tackle the \textit{multi-task pre-training} problem, enhancing sample efficiency and generalization ability for downstream tasks.
Building upon the success of temporal contrastive loss, exemplified by TACO~\citep{zheng2023taco}, in acquiring latent state representations that encapsulate individual task dynamics, our aim is to foster representation learning that effectively captures the intrinsic dynamics spanning a diverse set of tasks found in offline datasets.
Our overarching objective is to ensure that these learned representations exhibit the versatility to generalize across unseen tasks that share the underlying dynamic structures.

Nevertheless, when adapted for multitask offline pre-training, the online learning objective of TACO~\citep{zheng2023taco} poses a notable challenge. 
Specifically, TACO's mechanism, which utilizes the InfoNCE~\citep{oord2019representation} loss, categorizes all subsequent states $s_{t+k}$ in the batch as negative examples. While this methodology has proven effective in single-task reinforcement learning scenarios, it encounters difficulties when extended to a multitask context. During multitask offline pretraining, image observations within a batch can come from different tasks with vastly different visual appearances, rendering the contrastive InfoNCE loss significantly less effective.

\begin{figure}[!htbp]
\centering
\vspace{-1.0em}
\includegraphics[width=0.75\columnwidth]{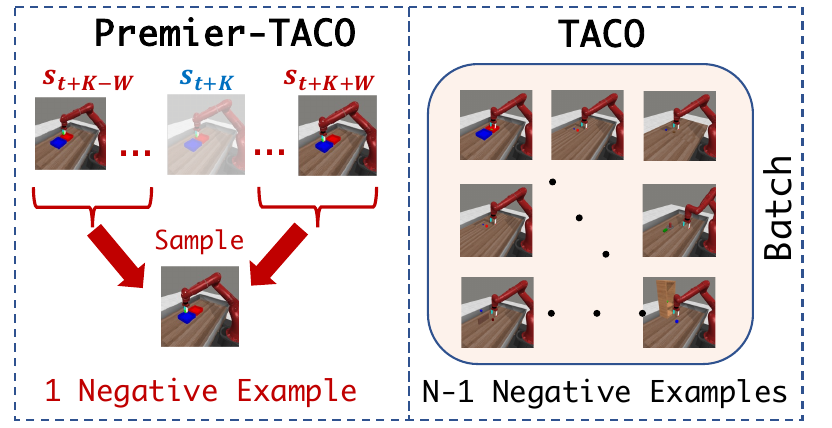}
\caption{Difference between \ours and TACO for sampling negative examples.} 
\label{fig:taco-vs-premier-taco}
\vspace{-1.0em}
\end{figure}

\textbf{Offline Pretraining Objective.}  We propose a straightforward yet highly effective mechanism for selecting challenging negative examples. Instead of treating all the remaining examples in the batch as negatives, \ours selects the negative example from a window centered at state $s_{t+k}$ within the same episode as shown in \Cref{fig:taco-vs-premier-taco}. 

This approach is both  computationally efficient and more statistically powerful due to negative examples which are challenging to distinguish from similar positive examples forcing the model capture temporal dynamics differentiating between positive and negative examples. 
In practice, this allows us to use much smaller batch sizes for \ours. On \mw, with only $\frac{1}{8}$ of the batch size (512 vs. 4096), \ours achieves a 25\% performance gain compared to TACO, saving around 87.5\% of computational time.

\begin{figure*}[!htbp]
\begin{centering}
\includegraphics[width=0.7\linewidth]{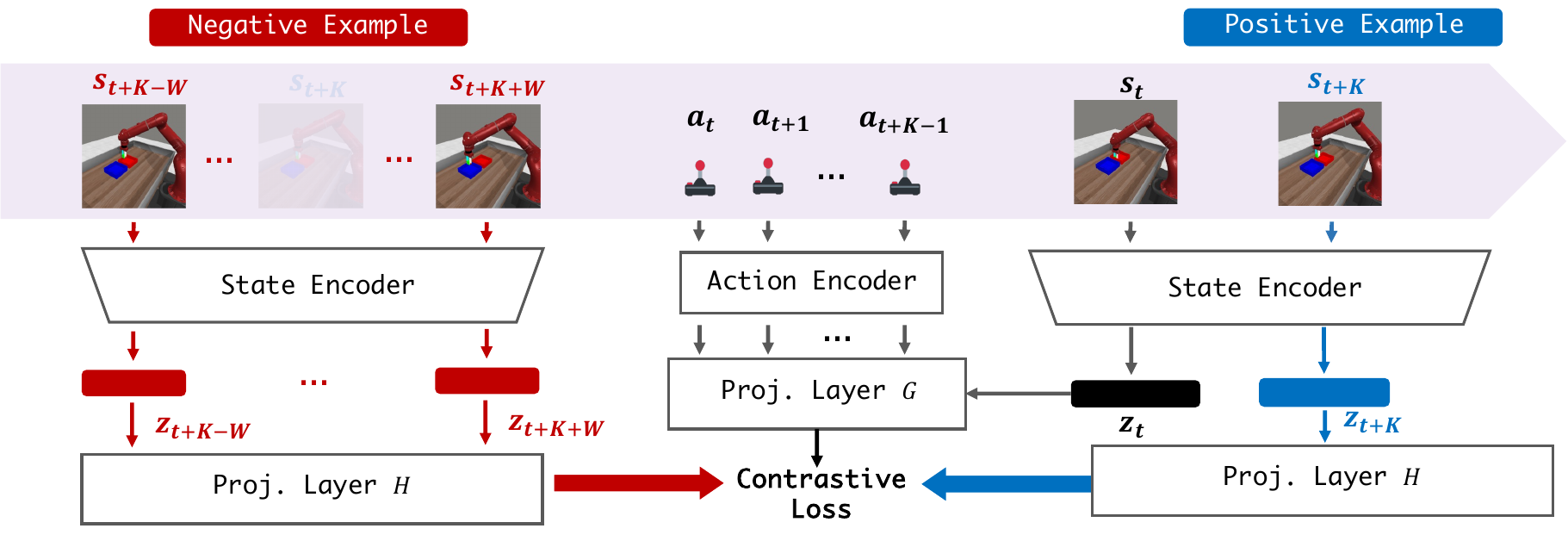} 
 \caption{An illustration of \ours contrastive loss design. The two `State Encoder's are identical, as are the two `Proj. Layer $H$'s. One negative example is sampled from the neighbors of framework $s_{t+K}$.}
 \label{fig:premier-taco-diagram}
  \end{centering}
 \end{figure*}

In \Cref{fig:premier-taco-diagram}, we illustrate the design of \ours objective.
Specifically, given a batch of state and action sequence transitions $\{(s_{t}^{(i)},[a^{(i)}_t,..., a^{(i)}_{t+K-1}],s^{(i)}_{t+K})\}_{i=1}^{N}$, let $z_t^{(i)}=\phi(s_t^{(i)})$, $u_t^{(i)}=\psi(a_t^{(i)})$ be latent state and latent action embeddings respectively.  
Furthermore, let $\widetilde{s_{t+K}^{(i)}}$ be a  negative example uniformly sampled from the window of size $W$ centered at $s_{t+K}$: $(s_{t+K-W},..., s_{t+K-1}, s_{t+K+1}, ..., s_{t+K+W})$ with $\widetilde{z_{t}^{(i)}}=\phi(\widetilde{s_t^{(i)}})$ a negative latent state.

Given these, define $g_t^{(i)}=G_\theta(z_{t}^{(i)},u^{(i)}_t,..., u^{(i)}_{t+K-1})$, $\widetilde{{h^{(i)}_t}}=H_\theta(\widetilde{z_{t+K}^{(i)}})$, and $h_t^{(i)}=H_\theta(z_{t+K}^{(i)})$ as embeddings of future predicted and actual latent states.  We optimize:
\vspace{-0.5em}
\resizebox{\columnwidth}{!}{$
\mathcal{J}_{\text{\ours}}(\phi, \psi, G_\theta, H_\theta)=-\frac{1}{N}\sum_{i=1}^{N} \log \frac{{g^{(i)}_t}^\top h^{(i)}_{t+K}}{{g^{(i)}_t}^\top h^{(i)}_{t+K} + \widetilde{{g^{(i)}_t}}^\top h^{(i)}_{t+K}} 
$}


\textbf{Few-shot Generalization.} 
After pretraining the representation encoder, we leverage our pretrained model $\Phi$ to learn policies for downstream tasks. To learn the policy $\pi$ with the state representation $\Phi(s_t)$ as inputs, we use behavior cloning (BC) with a few expert demonstrations. For different control domains, we employ significantly fewer demonstrations for unseen tasks than what is typically used in other baselines. This underscores the substantial advantages of \ours in few-shot generalization. More details about the experiments on downstream tasks will be provided in Section~\ref{sec:exp}. 

\vspace{-1em}
\section{Experiment}
\label{sec:exp}
\vspace{-0.5em}

\begin{figure*}[!htbp]
\centering
\includegraphics[width=0.85\linewidth]{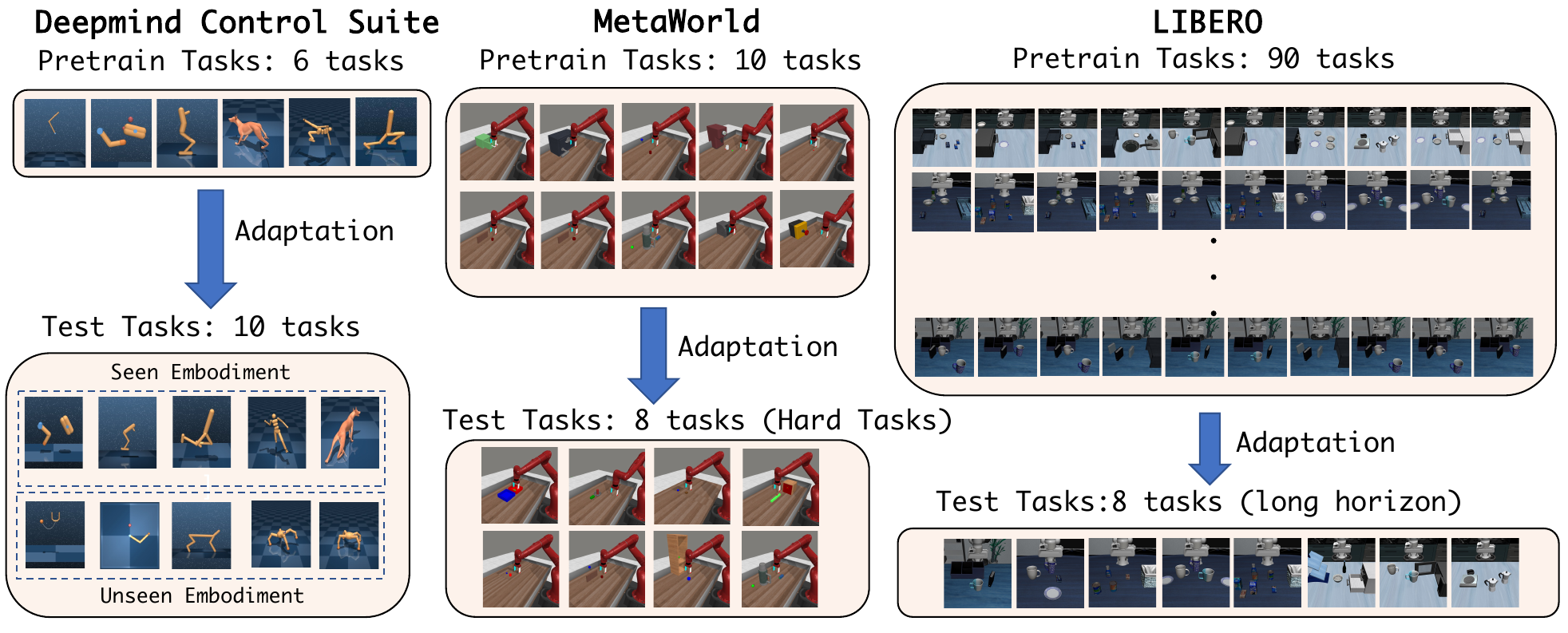}
\vspace{-10pt}
\caption{Pretrain and Test Tasks split for Deepmind Control Suite, MetaWorld and Libero. The left figures are Deepmind Control Suite tasks and the right figures MetaWorld tasks.}
\label{fig: pretrain tasks}
\end{figure*}

In our empirical evaluations, we consider three benchmarks, Deepmind Control Suite~\citep{dmc} for locomotion control, MetaWorld~\citep{metaworld} and LIBERO~\citep{libero} for robotic manipulation tasks. 
It is important to note the varied sources of data employed for pretraining in these benchmarks. 
For the Deepmind Control Suite, our pretraining dataset comes from the replay buffers of online reinforcement learning (RL) agents. 
In MetaWorld, the dataset is generated through a pre-defined scripted policy. 
In LIBERO, we utilize its provided demonstration dataset, which was collected through human teleoperation.
By evaluating on a wide range of pretraining data types that have been explored in previous works, we aim to provide a comprehensive evaluation for the pretraining effects of \ours.\looseness=-1

\textbf{Deepmind Control Suite (DMC)}: We consider a selection of 16 challenging tasks from Deepmind Control Suite. 
Note that compared with prior works such as~\cite{majumdar2023vc1, sun2023smart}, we consider much harder tasks, including ones from the humanoid and dog domains, which feature intricate kinematics, skinning weights and collision geometry. 
For pretraining, we select six tasks (\textbf{DMC-6}), including Acrobot Swingup, Finger Turn Hard, Hopper Stand, Walker Run, Humanoid Walk, and Dog Stand. 
We generate an exploratory dataset for each task by sampling trajectories generated in exploratory stages of a DrQ-v2~\citep{yarats2022drqv2} learning agent. 
In particular, we sample 1000 trajectories from the online replay buffer of DrQ-v2 once it reaches the convergence performance.
This ensures the diversity of the pretraining data, but in practice, such a high-quality dataset could be hard to obtain. 
So, later in the experiments, we will also relax this assumption and consider pretrained trajectories that are sampled from uniformly random actions.
In terms of the encoder architecture, we pretrain \ours with the same shallow ConvNet encoder as in DrQv2~\cite{yarats2022drqv2}.

\textbf{MetaWorld}: We select a set of 10 tasks for pretraining, which encompasses a variety of motion patterns of the Sawyer robotic arm and interaction with different objects. 
To collect an exploratory dataset for pretraining, we execute the scripted policy with Gaussian noise of a standard deviation of 0.3 added to the action.  
After adding such a noise, the success rate of collected policies on average is only around 20\% across ten pretrained tasks.
We use the same encoder network architecture as DMC.

\begin{table*}[!t]\small
\vspace{-0.5em}
\centering
\renewcommand{\arraystretch}{1.4}
\resizebox{\textwidth}{!}{%
\setlength{\tabcolsep}{3pt}
\begin{tabular}{p{1.9cm}<{\centering}  p{2.1cm}<{\centering} p{1.3cm}<{\centering} p{1.4cm}<{\centering} p{1.4cm}<{\centering} p{1.4cm}<{\centering}  p{1.3cm}<{\centering} p{1.3cm}<{\centering} p{1.3cm}<{\centering} p{1.3cm}<{\centering} p{1.3cm}<{\centering}>{\columncolor{gray!25}}p{2.3cm}<{\centering}}
\toprule
  & \textbf{DMControl}  &   \multicolumn{10}{c}{Models}   \\ 
   & \textbf{Tasks} & LfS & SMART & Best PVRs & TD3+BC &  Inverse & CURL & ATC  & SPR & TACO & \textbf{\ours}\\  
\hline
\multirow{5}{*}{\begin{tabular}[c]{@{}c@{}}\textbf{Seen}\\\textbf{Embodiments}\end{tabular}} 
 & Finger Spin     &  $34.8\pm 3.4$ & $44.2\pm 8.2$ & $38.4\pm 9.3$        & $68.8\pm 7.1$     & $33.4\pm8.4$    & $35.1\pm9.6$    & $51.1\pm 9.4$  & $55.9\pm 6.2$ & $28.4\pm 9.7$ & $\bm{75.2\pm 0.6}$  \\
 
 &  Hopper Hop     &  $8.0 \pm 1.3$ & $14.2\pm 3.9$ & $23.2 \pm 4.9$       & $49.1\pm 4.3$     & $48.3\pm5.2$    & $28.7\pm5.2$    & $34.9\pm 3.9$  & $52.3\pm7.8$ & $21.4\pm 3.4$ & $\bm{75.3\pm 4.6}$ \\
 
 &  Walker Walk    &  $30.4 \pm 2.9$& $54.1\pm 5.2$ & $32.6\pm 8.7$        & $65.8\pm 2.0$     & $64.4\pm5.6$    & $37.3\pm 7.9$   & $44.6\pm 5.0$  & $72.9\pm 1.5$ & $30.6\pm 6.1$ & $\bm{88.0\pm 0.8}$ \\
 
 &  Humanoid Walk  &  $15.1\pm 1.3$ & $18.4 \pm 3.9$& $30.1\pm 7.5$        & $34.9\pm8.5$      & $41.9\pm 8.4$   & $19.4\pm 2.8$   &  $35.1\pm 3.1$ & $30.1\pm6.2$ & $29.1\pm 8.1$ & $\bm{51.4\pm 4.9}$ \\
 
  & Dog Trot       &  $52.7\pm 3.5$ & $59.7\pm 5.2$ & $73.5\pm 6.4$       & $82.3\pm 4.4$    & $85.3\pm2.1$     & $71.9\pm 2.2$   &  $84.3\pm0.5$  & $79.9\pm 3.8$& $80.1\pm 4.1$ & $\bm{93.9 \pm 5.4}$ \\
\hline
\multirow{5}{*}{\begin{tabular}[c]{@{}c@{}}\textbf{Unseen}\\\textbf{Embodiments}\end{tabular}} 
 & Cup Catch        & $56.8 \pm 5.6$& $66.8\pm 6.2$ &  $93.7\pm1.8$                           & $97.1\pm1.7$           & $96.7\pm 2.6$          & $96.7\pm2.6$   & $96.2\pm 1.4$           & $96.9\pm 3.1$ & $88.7\pm 3.2$ & $\bm{98.9 \pm 0.1}$  \\
 & Reacher Hard     & $34.6\pm 4.1$ & $52.1\pm 3.8$ &  $64.9\pm5.8$                           & $59.6\pm9.9$           & $61.7\pm 4.6$          & $50.4\pm 4.6$  & $56.9\pm 9.8$ & $62.5\pm7.8$ & $58.3\pm 6.4$ & $\bm{81.3\pm 1.8}$ \\
 & Cheetah Run      & $25.1\pm 2.9$ & $41.1\pm 7.2$ &  $39.5\pm 9.7$                          & $50.9\pm 2.6$          & $51.5\pm 5.5$          & $36.8\pm 5.4$  & $30.1\pm 1.0$  & $40.2\pm 9.6$& $23.2\pm 3.3$ & $\bm{65.7 \pm 1.1}$  \\
&  Quadruped Walk   & $61.1\pm 5.7$ & $45.4\pm 4.3$ &  $63.2\pm 4.0$ & $76.6\pm 7.4$          & $82.4\pm 6.7$          & $72.8\pm 8.9$          & $81.9\pm 5.6$ & $65.6\pm 4.0$ & $63.9\pm 9.3$ & $\bm{83.2\pm 5.7}$  \\
&  Quadruped Run    & $45.0\pm 2.9$ & $27.9\pm 5.3$ &  $64.0\pm 2.4$ & $48.2\pm 5.2$          & $52.1\pm1.8$           & $55.1\pm 5.4$          & $2.6\pm 3.6$  & $68.2\pm 3.2$& $50.8\pm 5.7$ & $\bm{76.8\pm7.5}$ \\
\hline
\multicolumn{2}{c}{\textbf{Mean Performance}} & $38.2$& $42.9$ &  $52.3$                           & $63.3$           & $61.7$         & $50.4$   & $52.7$            &$62.4$ & $47.5$ & $\bm{79.0}$  \\  
\bottomrule
\end{tabular}}
\caption{\textbf{[(W1) Versatility (W2) Efficiency]} \textbf{Few-shot Behavior Cloning (BC) for unseen task of DMC.} Performance (Agent Reward / Expert Reward) of baselines and \ours on 10 unseen tasks on \dmc. \textbf{Bold} numbers indicate the best results.
Agent Policies are evaluated every 1000 gradient steps for a total of 100000 gradient steps and we report the average performance over the 3 best epochs over the course of learning.
\ours outperforms all the baselines, showcasing its superior efficacy in generalizing to unseen tasks with seen or \textbf{unseen embodiments}.}
\label{tab:dmc_unseen_task}
\vspace{-1em}
\end{table*}


\begin{table*}[!t]\small
\centering
\renewcommand{\arraystretch}{1.4}
\resizebox{\textwidth}{!}{%
\setlength{\tabcolsep}{3pt}

\begin{tabular}{ p{2.3cm}<{\centering} p{1.7cm}<{\centering} p{1.6cm}<{\centering} p{1.6cm}<{\centering} p{1.6cm}<{\centering}  p{1.6cm}<{\centering} p{1.6cm}<{\centering} p{1.6cm}<{\centering} p{1.6cm}<{\centering} p{1.6cm}<{\centering} >{\columncolor{gray!25}}p{2.3cm}<{\centering}}
\toprule
\textbf{\mw}  &   \multicolumn{10}{c}{Models}   \\ 
   \textbf{Unseen Tasks} & LfS & SMART & Best PVRs & TD3+BC &  Inverse & CURL & ATC & SPR & TACO  & \textbf{\ours}\\  
\hline
Bin Picking    &  $62.5 \pm 12.5$ & $71.3\pm 9.6$ &   $60.2 \pm 4.3$    &  $50.6\pm 3.7$    &  $55.0\pm 7.9$ &   $45.6\pm 5.6$ & $55.6 \pm 7.8$ & $67.9\pm 6.4$ & $67.3\pm 7.5$ & $\bm{78.5 \pm 7.2}$ \\

Disassemble    & $56.3 \pm 6.5$  & $52.9\pm 4.5$ &   $70.4 \pm 8.9 $   &   $56.9\pm 11.5$   &  $53.8\pm 8.1$  &    $66.2 \pm 8.3$ &   $45.6 \pm 9.8$ & $48.8\pm 5.4$& $51.3\pm 10.8$ & $\bm{86.7 \pm 8.9}$  \\
Hand Insert    & $34.7 \pm 7.5$  & $34.1\pm 5.2$&  $ 35.5 \pm 2.3 $    &  $46.2\pm5.2$    &    $50.0\pm 3.5$ &    $49.4\pm 7.6$&   $51.2\pm 1.3$  & $52.4\pm 5.2$& $56.8\pm 4.2$ & $\bm{75.0 \pm 7.1}$  \\
Peg Insert Side   & $28.7 \pm 2.0 $ & $20.9\pm 3.6$& $48.2 \pm 3.6 $      &  $30.0\pm 6.1$    &  $33.1\pm 6.2$  &    $28.1 \pm 3.7$&   $31.8\pm 4.8$ & $39.2\pm 7.4$ & $36.3\pm4.5$ & $\bm{62.7 \pm 4.7}$  \\
Pick Out Of Hole      & $53.7 \pm 6.7$  & $65.9\pm 7.8$&  $66.3 \pm 7.2$    &  $46.9\pm 7.4$    &   $50.6\pm 5.1$ &    $43.1\pm 6.2$ &   $54.4 \pm 8.5$ & $55.3\pm 6.8$ & $52.9\pm7.3$ & $\bm{72.7 \pm 7.3}$  \\
Pick Place Wall       & $40.5 \pm 4.5$  & $62.8\pm 5.9$&  $63.2 \pm 9.8 $     &   $63.8\pm 12.4$   &   $71.3\pm 11.3$ &    $73.8\pm 11.9$&   $68.7\pm 5.5$& $72.3\pm 7.5$ & $37.8\pm 8.5$ & $\bm{80.2 \pm 8.2}$  \\
Shelf Place    & $26.3 \pm 4.1$  & $57.9 \pm 4.5$&  $ 32.4 \pm 6.5   $  &    $45.0\pm 7.7$  &$36.9\pm 6.7$ & $35.0\pm 10.8$ &   $35.6\pm 10.7$&  $38.0\pm 6.5$& $25.8\pm 5.0$ & $\bm{70.4 \pm 8.1}$ \\
Stick Pull  & $46.3 \pm 7.2 $ & $65.8\pm 8.2$ &  $52.4 \pm 5.6$   &    $72.3\pm 11.9$  &    $57.5\pm 9.5$ &    $43.1 \pm 15.2$&    $72.5\pm 8.9$ & $68.5\pm 9.4$  & $52.0\pm 10.5$ & $\bm{80.0 \pm 8.1}$ \\
\hline
\textbf{Mean} & $43.6$ & $53.9$ &  $53.6$   &    $51.5$  &    $51.0$ &    $48.3$&    $51.9$ & $55.3$ & $47.5$ & $\bm{75.8}$ \\
\bottomrule
\end{tabular}}
\caption{\textbf{[(W1) Versatility (W2) Efficiency]} \textbf{Five-shot Behavior Cloning (BC) for unseen task of \mw.} Success rate of \ours and baselines across 8 hard unseen tasks on \mw. Results are aggregated over 4 random seeds. \textbf{Bold} numbers indicate the best results. }
\label{tab:mw_unseen_task}
\vspace{-1em}
\end{table*}

\textbf{LIBERO}: We pretrain on 90 short-horizon manipulation tasks (\textbf{LIBERO-90}) with human demonstration dataset provided by the original paper. 
For each task, it contains 50 trajectories of human teleoperated trajectories.
We use ResNet18 encoder~\cite{resnet} to encode the image observations of resolution $128\times 128$.
For the downstream task, we assess the few-shot imitation learning performance on the first 8 long-horizon tasks of LIBERO-LONG.

\textbf{Baselines.} We compare \ours with the following representation pretraining baselines:
\begin{list}{$\rhd$}{\topsep=0.ex \leftmargin=0.1in \rightmargin=0.in \itemsep =0.0in}
\item Learn from Scratch: Behavior Cloning with randomly initialized shallow ConvNet encoder. We carefully implement the behavior cloning from scratch baseline. 
For DMC and \mw, following~\cite{hansen2022on}, we include the random shift data augmentation into behavior cloning.
For LIBERO, we take the ResNet-T model architecture in~\citep{libero}, which uses a transformer decoder module on top of the ResNet encoding to extract temporal information from a sequence of observations, addressing the non-Markovian characteristics inherent in human demonstration. 
\begin{figure*}[t!]
\centering
\includegraphics[width=0.6\linewidth]{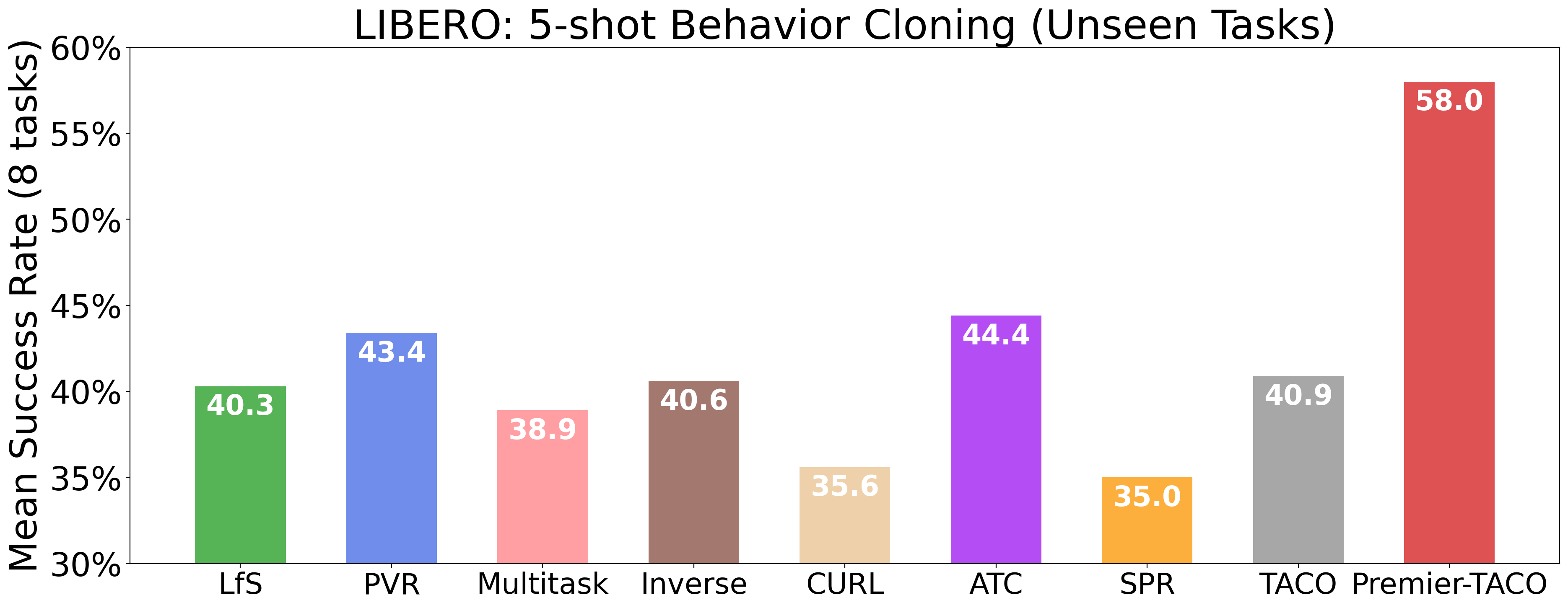} 
\vspace{-1.0em}
 \caption{\textbf{[(W1) Versatility (W2) Efficiency]} Mean success rate of 5-shot imitation learning for 8 unseen tasks in LIBERO. Results are aggregated over 4 random seeds. \textbf{Bold} numbers indicate the best results. See the results for individual tasks in Table~\ref{tab:libero_unseen_task}.}
  \label{fig:libero_unseen}
  \vspace{-1.0em}
 \end{figure*}
\item Policy Pretraining: We first train a multitask policy by TD3+BC~\citep{scott21td3bc} on the pretraining dataset. While numerous alternative offline RL algorithms exist, we choose TD3+BC as a representative due to its simplicity and great empirical performance.
For LIBERO, we use Multitask BC since offline RL in generally does not perform well on the imitation learning benchmark with human demonstrated dataset.
After pretraining, we take the pretrained ConvNet encoder and drop the policy MLP layers. 
\item Pretrained Visual Representations (PVRs): We evaluate the state-of-the-art frozen pretrained visual representations including PVR~\citep{parisi22pvr}, MVP~\citep{xiao2022mvp}, R3M~\citep{nair2022r3m} and VC-1~\citep{majumdar2023vc1}, and report the best performance of these PVRs models for each task. \looseness=-1
\item Control Transformer: SMART~\citep{sun2023smart} is a self-supervised representation pretraining framework which utilizes a maksed prediction objective for pretraining representation under Decision Transformer architecture, and then use the pretrained representation to learn policies for downstream tasks.
\item Inverse Dynamics Model: We pretrain an inverse dynamics model to predict actions and use the pretrained representation for downstream task.
\item Contrastive/Self-supervised Learning Objectives: CURL~\citep{laskin20curl}, ATC~\citep{adam21atc}, SPR~\citep{schwarzer2021dataefficient, schwarzer2021pretraining}. CURL and ATC are two approaches that apply contrastive learning into sequential decision making problems. 
While CURL treats augmented states as positive pairs, it neglects the temporal dependency of MDP.
In comparison, ATC takes the temporal structure into consideration. The positive example of ATC is an augmented view of a temporally nearby state.  
SPR applies BYOL objecive~\citep{grill20byol} into sequential decision making problems by pretraining state representations that are self-predictive of future states.
\end{list}
\textbf{Pretrained feature representation by \ours facilitates effective few-shot adaptation to unseen tasks.}  We measure the performance of pretrained visual representation for few-shot imitation learning of unseen downstream tasks in both DMC and MetaWorld. 
In particular, for DMC, we use \textbf{20 expert trajectories} for imitation learning except for the two hardest tasks, Humanoid Walk and Dog Trot, for which we use 100 trajectories instead. 
Note that we only use $\frac{1}{5}$ of the number of expert trajectories used in \cite{majumdar2023vc1} and $\frac{1}{10}$ of those used in \cite{sun2023smart}.

We record the performance of the agent by calculating the ratio of $\displaystyle{\frac{\text{Agent Reward}}{\text{Expert Reward}}}$, where Expert Reward is the episode reward of the expert policy used to collect demonstration trajectories. 
For \mw and LIBERO, we use \textbf{5 expert trajectories} for all downstream tasks, and we use task success rate as the performance metric. 
In Table~\ref{tab:dmc_unseen_task} Table~\ref{tab:mw_unseen_task}, and~\Cref{fig:libero_unseen} we present the results for \dmc, \mw, and LIBERO, respectively. 
As shown here, pretrained representation of \ours significantly improves the few-shot imitation learning performance compared with \lfs, with a \textbf{101\%} improvement on \dmc and \textbf{74\%} improvement on \mw, respectively. Moreover, it also outperforms all the baselines across all tasks by a large margin.
In LIBERO, consistent with what is observed in ~\citep{libero}, existing pretraining methods on large-scale multitask offline dataset fail to enhance downstream policy learning performance. 
In particular, methods like multitask pretraining actually degrade downstream policy learning performance. 
In contrast, using ResNet-18 encoders pretrained by \ours significantly boosts few-shot imitation learning performance by a substantial margin.

\begin{figure*}[t!]
\centering
\includegraphics[width=0.7\linewidth]{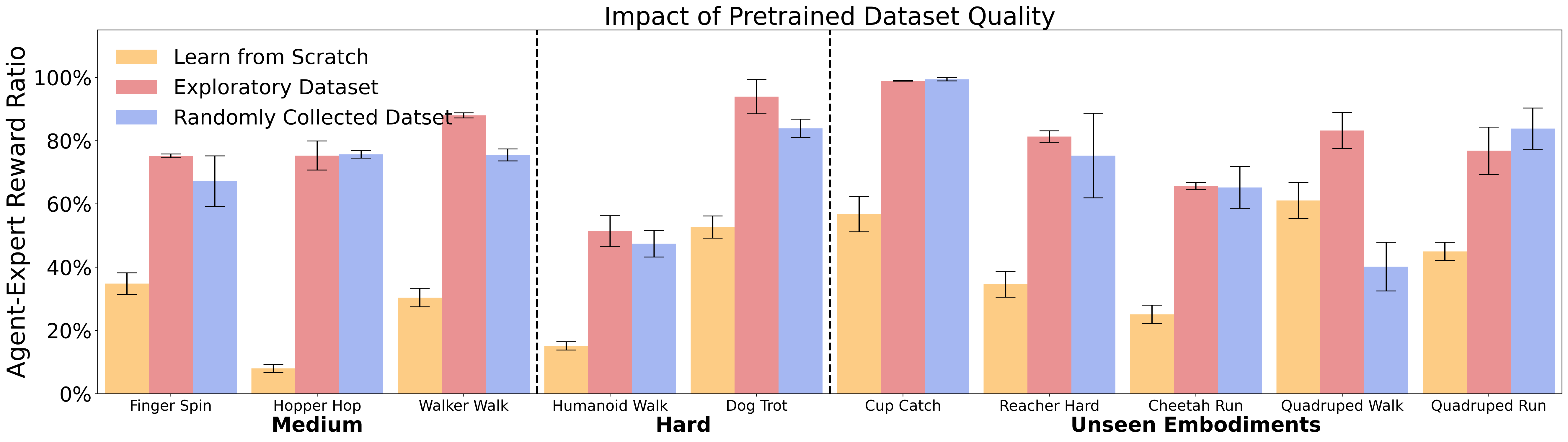} 
\vspace{-0.5em}
 \caption{\textbf{[(W3) Robustness]} \ours pretrained with exploratory dataset vs. \ours pretrained with randomly collected dataset}
  \label{low_quality}
  \vspace{-1.5em}
 \end{figure*}
\begin{figure*}[htbp!]
\begin{centering}
\includegraphics[width=0.7\linewidth]{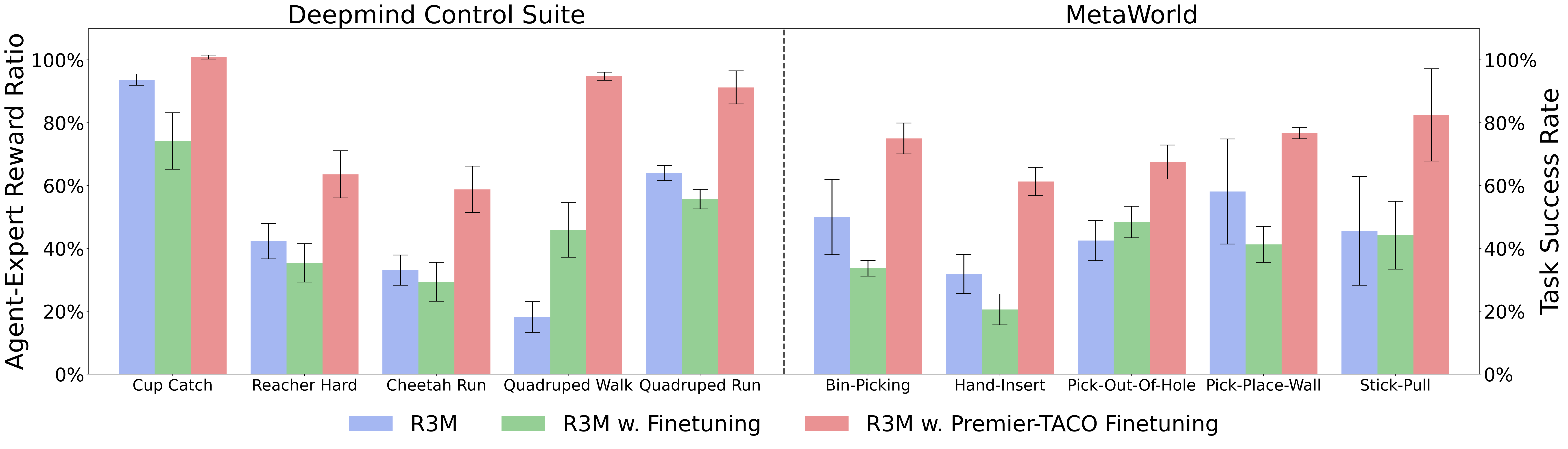} 
\vspace{-1em}
 \caption{\textbf{[(W4) Compatibility]} Finetune R3M~\citep{nair2022r3m}, a generalized Pretrained Visual Encoder with \ours learning objective vs. R3M with in-domain finetuning in \dmc and \mw.}\label{fig:r3m_finetune}
 \vspace{-1.0em}
  \end{centering}
 \end{figure*}
 
\begin{figure}[h!]
\vspace{-0.5em}
\includegraphics[width=\columnwidth]{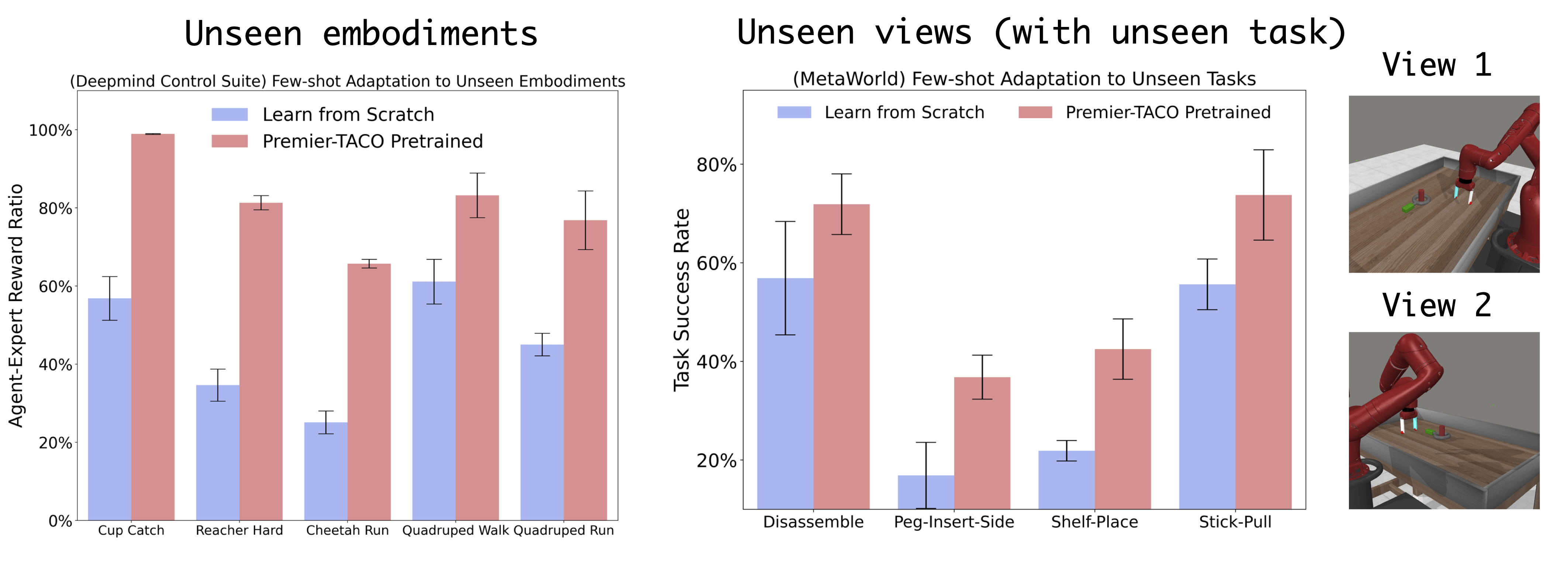}
    \centering
    \vspace{-2.0em}
    \caption{\textbf{[(W1) Versatility]} (\textbf{Left}) DMC: Generalization of \ours pre-trained visual representation to unseen embodiments. (\textbf{Right}) \mw: Few-shot adaptation to unseen tasks from an unseen camera view}
    \vspace{-2.0em}
    \label{fig:unseen_embodiment_views}
\end{figure}

\textbf{\ours pre-trained representation enables knowledge sharing across different embodiments.} 
Ideally, a resilient and generalizable state feature representation ought not only to encapsulate universally applicable features for a given embodiment across a variety of tasks, but also to exhibit the capability to generalize across distinct embodiments. 
Here, we evaluate the few-shot behavior cloning performance of \ours pre-trained encoder from $\textbf{DMC-6}$ on four tasks featuring unseen embodiments: Cup Catch, Cheetah Run, and Quadruped Walk. In comparison to \lfs, as shown in~\Cref{fig:unseen_embodiment_views} (left), \ours pre-trained representation realizes an \textbf{82\%} performance gain, demonstrating the robust generalizability of our pre-trained feature representations. \looseness=-1

\textbf{\ours Pretrained Representation is also generalizable to unseen tasks with camera views.} Beyond generalizing to unseen embodiments, an ideal robust visual representation should possess the capacity to adapt to unfamiliar tasks under novel camera views. 
In~\Cref{fig:unseen_embodiment_views} (right), we evaluate the five-shot learning performance of our model on four previously unseen tasks in MetaWorld with a new view. 
In particular, during pretraining, the data from MetaWorld are generated using the same view as employed in ~\citep{nicklas22tdmpc, seo22mwm}. Then for downstream policy learning, the agent is given five expert trajectories under a different corner camera view, as depicted in the figure. 
Notably, \ours also achieves a substantial performance enhancement, thereby underscoring the robust generalizability of our pretrained visual representation.

\textbf{\ours Pre-trained Representation is resilient to low-quality data.} We evaluate the resilience of \ours by employing randomly collected trajectory data from \dmc for pretraining and compare it with \ours representations pretrained using an exploratory dataset and the learn-from-scratch approach. As illustrated in Figure~\ref{low_quality}, across all downstream tasks, even when using randomly pretrained data, the \ours pretrained model still maintains a significant advantage over learning-from-scratch. When compared with representations pretrained using exploratory data, there are only small disparities in a few individual tasks, while they remain comparable in most other tasks. This strongly indicates the robustness of \ours to low-quality data. Even without the use of expert control data, our method is capable of extracting valuable information.

\textbf{Pretrained visual encoder finetuning with \ours. }
In addition to evaluating our pretrained representations across various downstream scenarios, we also conducted fine-tuning on pretrained visual representations using in-domain control trajectories following \ours framework. Importantly, our findings deviate from the observations made in prior works like \citep{hansen2022on} and \citep{majumdar2023vc1}, where fine-tuning of R3M~\citep{nair2022r3m} on in-domain demonstration data using the task-centric behavior cloning objective, resulted in performance degradation. We speculate that two main factors contribute to this phenomenon. First, a domain gap exists between out-of-domain pretraining data and in-domain fine-tuning data. Second, fine-tuning with few-shot learning can lead to overfitting for large pretrained models.
 
To further validate the effectiveness of our \ours approach, we compared the results of R3M with no fine-tuning, in-domain fine-tuning~\citep{hansen2022on}, and fine-tuning using our method on selected \dmc and \mw pretraining tasks. Figure~\ref{fig:r3m_finetune} unequivocally demonstrate that direct fine-tuning on in-domain tasks leads to a performance decline across multiple tasks. However, leveraging the \ours learning objective for fine-tuning substantially enhances the performance of R3M. This not only underscores the role of our method in bridging the domain gap and capturing essential control features but also highlights its robust generalization capabilities. 
Furthermore, these findings strongly suggest that our \ours approach is highly adaptable to a wide range of multi-task pretraining scenarios, irrespective of the model's size or the size of the pretrained data.

\textbf{Ablation Study - Batch Size}: Compared with TACO, the negative example sampling strategy employed in \ours allows us to sample harder negative examples within the same episode as the positive example. 
This implies the promising potential to significantly improve the performance of existing pretrained models across diverse domains. The full results of finetuning on all 18 tasks including \dmc and \mw are in Appendix~\ref{app:finetune}.
We expect \ours to work much better with small batch sizes, compared with TACO where the negative examples from a given batch could be coming from various tasks  and thus the batch size required would scale up linearly with the number of pretraining tasks.  
In ours previous experimental results, \ours is pretrained with a batch size of 4096, a standard batch size used in contrastive learning literature. 
Here, to empirically verify the effects of different choices of the pretraining batch size, we train \ours and TACO with different batch sizes and compare their few-shot imitation learning performance.  \looseness=-1

\begin{figure}[h!]
\includegraphics[width=\columnwidth]{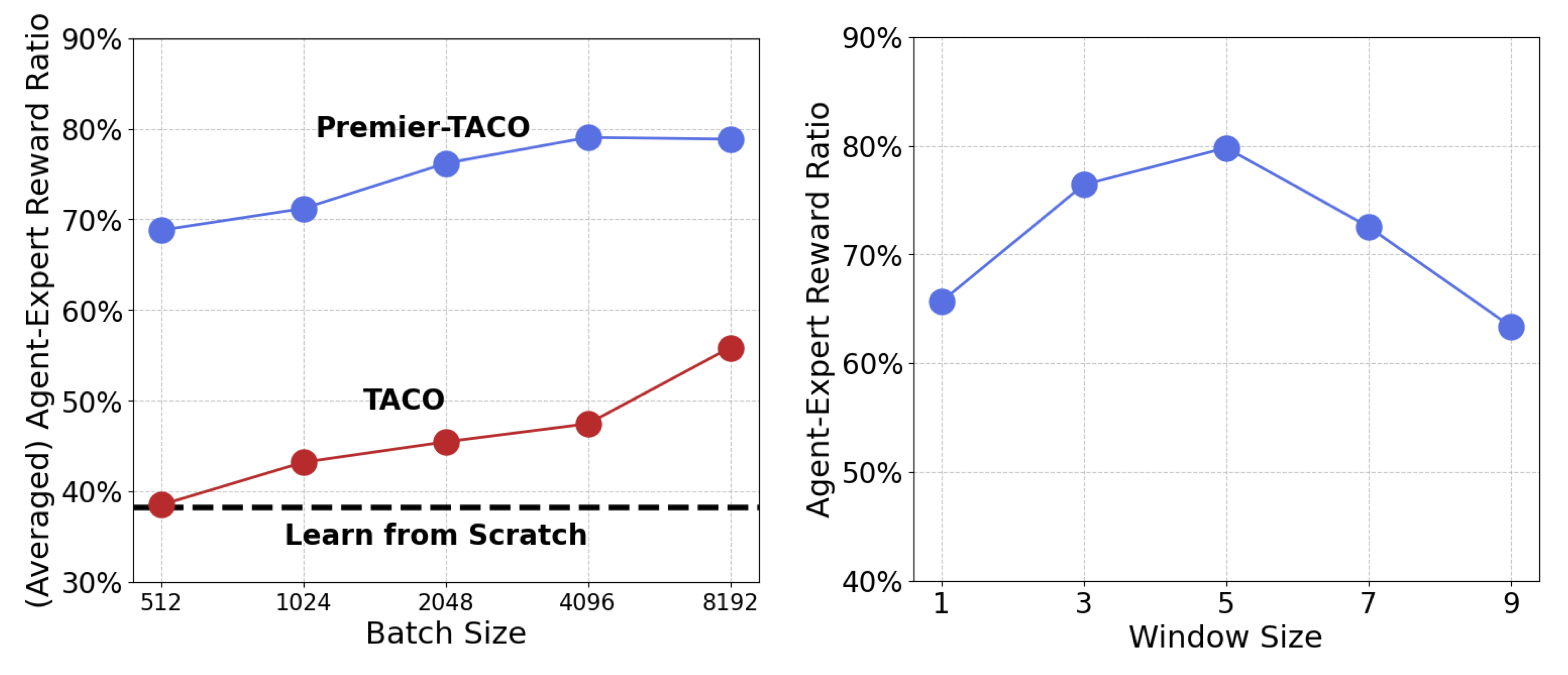}
    \centering
    \vspace{-2.0em}
    \caption{\textbf{[(W1) Versatility]} (\textbf{Left}) \ours vs. TACO on 10 Deepmind Control Suite Tasks across different batch sizes. (\textbf{Right}) Averaged performance of \ours on 10 Deepmind Control Suite Tasks across different window sizes}
    \label{fig:batch_size_window_size}
    \vspace{-1.0em}
\end{figure}

Figure~\ref{fig:batch_size_window_size} (left) displays the average performance of few-shot imitation learning across all ten tasks in the DeepMind Control Suite. 
As depicted in the figure, our model significantly outperform TACO across all batch sizes tested in the experiments, and exhibits performance saturation beyond a batch size of 4096. This observation substantiate that the negative example sampling strategy employed by \ours is indeed the key for the success of multitask offline pretraining.

\textbf{Ablation Study - Window Size}: In \ours, the window size $W$ determines the hardness of the negative example. 
A smaller window size results in negative examples that are more challenging to distinguish from positive examples, though they may become excessively difficult to differentiate in the latent space. 
Conversely, a larger window size makes distinguishing relatively straightforward, thereby mitigating the impacts of negative sampling. 
In preceding experiments, a consistent window size of 5 was applied across all trials on both the DeepMind Control Suite and MetaWorld.  
Here in~\Cref{fig:batch_size_window_size} (right) we empirically evaluate the effects of varying window sizes on the average performance of our model across ten DeepMind Control Tasks. 
Notably, our observations reveal that performance is comparable when the window size is set to 3, 5, or 7, whereas excessively small ($W=1$) or large ($W=9$) window sizes lead to worse performance.
\section{Related Work}
\label{sec:relate}

Existing works, including R3M~\citep{nair2022r3m}, VIP~\citep{ma2023vip}, MVP~\citep{xiao2022mvp}, PIE-G~\citep{neurips2022yuan}, and VC-1~\citep{majumdar2023vc1}, focus on self-supervised pre-training for building foundation models but struggle with the domain gap in sequential decision-making tasks. 
Recent studies, such as one by \citet{hansen2022on}, indicate that models trained from scratch often outperform pre-trained representations. 
Approaches like SMART~\citep{sun2023smart} and DualMind~\citep{wei2023imitation} offer control-centric pre-training, but at the cost of extensive fine-tuning or task sets. 
Contrastive learning techniques like CURL~\citep{laskin20curl}, CPC~\citep{henaff20cpcv2}, ST-DIM~\citep{ankesh2019stdim}, and ATC~\citep{adam21atc} have succeeded in visual RL, but mainly focus on high-level features and temporal dynamics without a holistic consideration of state-action interactions, a gap partially filled by TACO~\citep{zheng2023taco}. 
Our work builds upon these efforts but eliminates the need for extensive task sets and fine-tuning, efficiently capturing control-relevant features. 
This positions our method as a distinct advancement over DRIML~\citep{bogdan20driml} and Homer~\citep{Homer}, which require more computational or empirical resources.

A detailed discussion of related work is in Appendix~\ref{sec:relate_apd}.

\section{Conclusion}
\label{sec:con}
This paper introduces \ours, a robust and highly generalizable representation pretraining framework for few-shot policy learning. We propose a temporal contrastive learning objective that excels in multi-task representation learning during the pretraining phase, thanks to its efficient negative example sampling strategy. Extensive empirical evaluations spanning diverse domains and tasks underscore the remarkable effectiveness and adaptability of \ours's pre-trained visual representations to unseen tasks, even when confronted with unseen embodiments, different views, and data imperfections. Furthermore, we demonstrate the versatility of \ours by showcasing its ability to fine-tune large pretrained visual representations like R3M~\citep{nair2022r3m} with domain-specific data, underscoring its potential for broader applications. \looseness=-1

\section*{Acknowledgements} 
Zheng, Wang, and Huang are supported by National Science Foundation NSF-IIS-2147276 FAI, DOD-ONR-Office of Naval Research under award number N00014-22-1-2335, DOD-AFOSR-Air Force Office of Scientific Research under award number FA9550-23-1-0048, DOD-DARPA-Defense Advanced Research Projects Agency Guaranteeing AI Robustness against Deception (GARD) HR00112020007, Adobe, Capital One and JP Morgan faculty fellowships.

\section*{Impact Statement}
This paper presents work whose goal is to advance the field of Machine Learning. There are many potential societal consequences of our work, none which we feel must be specifically highlighted here.

\bibliography{references}
\bibliographystyle{icml2024}

\appendix
\newpage
\onecolumn
\section{Detailed Discussion of Related Work}
\label{sec:relate_apd}
\textbf{Pretraining Visual Representations.} Existing works apply self-supervised pre-training from rich vision data to build foundation models. However, applying this approach to sequential decision-making tasks is challenging. Recent works have explored large-scale pre-training with offline data in the context of reinforcement learning. Efforts such as R3M~\citep{nair2022r3m}, VIP~\citep{ma2023vip}, MVP~\citep{xiao2022mvp}, PIE-G~\citep{neurips2022yuan}, and VC-1~\citep{majumdar2023vc1} highlight this direction. However, there's a notable gap between the datasets used for pre-training and the actual downstream tasks. In fact, a recent study~\citep{hansen2022on} found that models trained from scratch can often perform better than those using pre-trained representations, suggesting the limitation of these approachs.  It's important to acknowledge that these pre-trained representations are not control-relevant, and they lack explicit learning of a latent world model. In contrast to these prior approaches, our pretrained representations learn to capture the control-relevant features with an effective temporal contrastive learning objective.

For control tasks, several pretraining frameworks have emerged to model state-action interactions from high-dimensional observations by leveraging causal attention mechanisms. SMART~\citep{sun2023smart} introduces a self-supervised and control-centric objective to train transformer-based models for multitask decision-making, although it requires additional fine-tuning with large number of demonstrations during downstream time. As an improvement, DualMind~\citep{wei2023imitation} pretrains representations using 45 tasks for general-purpose decision-making without task-specific fine-tuning. 
Besides, some methods~\citep{{sekar2020planning}, {mendonca2021discovering}, {yarats2021reinforcement}, {sun2022transfer}} first learn a general representation by exploring the environment online, and then use this representation to train the policy on downstream tasks.
In comparison, our approach is notably more efficient and doesn't require training with such an extensive task set. Nevertheless, we provide empirical evidence demonstrating that our method can effectively handle multi-task pretraining.

\textbf{Contrastive Representation for Visual RL}
Visual RL~\cite{yarats2021drq, yarats2022drqv2,Hafner2020Dream, nicklas22tdmpc,xu2024drm, ji2024ace} is long-standing challenge due to the entangled problem of representation learning and credit assignment.
In the context of visual reinforcement learning (RL), contrastive learning plays a pivotal role in training robust state representations from raw visual inputs, thereby enhancing sample efficiency. CURL~\citep{laskin20curl} extracts high-level features by utilizing InfoNCE\citep{oord2019representation} to maximize agreement between augmented observations, although it does not explicitly consider temporal relationships between states. Several approaches, such as CPC~\citep{henaff20cpcv2}, ST-DIM~\citep{ankesh2019stdim}, and ATC~\citep{adam21atc} , introduce temporal dynamics into the contrastive loss. They do so by maximizing mutual information between states with short temporal intervals, facilitating the capture of temporal dependencies. DRIML~\citep{bogdan20driml} proposes a policy-dependent auxiliary objective that enhances agreement between representations of consecutive states, specifically considering the first action of the action sequence. Recent advancements by \citet{kim2022action, zhang2021learning} incorporate actions into the contrastive loss, emphasizing behavioral similarity. TACO~\citep{zheng2023taco} takes a step further by learning both state and action representations. It optimizes the mutual information between the representations of current states paired with action sequences and the representations of corresponding future states. In our approach, we build upon the efficient extension of TACO, harnessing the full potential of state and action representations for downstream tasks.  On the theory side, the Homer algorithm~\citep{Homer} uses a binary temporal contrastive objective reminiscent of the approach used here, which differs by abstracting actions as well states, using an ancillary embedding, removing leveling from the construction, and of course extensive empirical validation.

\textbf{Hard Negative Sampling Strategy in Contrastive Learning} Our proposed negative example sampling strategy in \ours is closely related to hard negative example mining in the literature of self-supervised learning as well as other areas of machine learning. Hard negative mining is indeed used in a variety of tasks, such as facial recognition \citep{wan2016bootstrapping}, object detection~\citep{abhinav2016}, tracking~\citep{nam2016}, and image-text retrieval~\citep{pang2019libra, li2021align}, by introducing negative examples that are more difficult than randomly chosen ones to improve the performance of models.
Within the regime of self-supervised learning, different negative example sampling strategies have also been discussed both empirically and theoretically to improve the quality of pretrained representation. 
In particular, \cite{robinson2021contrastive} modifies the original NCE objective by developing a distribution over negative examples, which prioritizes pairs with currently similar representations. \cite{kaltantidis2020} suggests to mix hard negative examples within the latent space. \cite{ma2021active} introduce a method to actively sample uncertain negatives by calculating the gradients of the loss function relative to the model’s most confident predictions. Furthermore, \cite{tabassum2022hard} samples negatives that combine the objectives of identifying model-uncertain negatives, selecting negatives close to the anchor point in the latent embedding space, and ensuring representativeness within the sample population.

\textbf{Comparison with Offline Meta-RL Methods} Compared with offline meta-rl methods, feature representation learning with self-supervised/contrastive objectives, such as Premier-TACO, can efficiently leverage low-quality datasets (e.g., datasets collected by rolling out random actions in the DeepMind Control Suite)~\cite{xu2023hdt, gao2023csro, mitchell2021offline}.
Offline meta RL, in contrast, relies on datasets with good coverage to learn effective policies and typically addresses tasks with smaller shifts between meta-training and meta-testing (e.g., varying velocities in MuJoCo's halfcheetah).

In particular, HDT~\cite{xu2023hdt} utilizes a hyper-network as an adaptation module to encode expert demonstrations and augment the base Decision Transformer~\cite{chen2021dt} model.
Unlike HDT, \ours can adapt to unseen embodiments with different action spaces by initializing a new policy head. In contrast, HDT's hyper-network architecture does not easily accommodate unseen action spaces without significant modifications.
CSRO~\cite{gao2023csro} focuses on smaller task distribution shift as in prior offline meta RL works, such as humanoid in MuJoCo running in different directions. 
The task representation learned in CSRO is also not able to handle the unseen action spaces. On the contrary, \ours tackles a broader problem and experimental setting, enabling our representation to generalize to unseen downstream tasks and unseen embodiments with new action spaces.

\textbf{Other pretraining schemes for decision-making.} In this work, we primarily focuses on pretraining visual state representations. Several other existing works aim to solve multitask pretraining for sequential decision making from a different perspective, through the discovery of temporal action abstractions (i.e. skills or options).
These works propose to pretrain temporally extended action primitives and subsequently use them for shortening the effective decision-making horizon during high-level policy induction, including CompILE~\citep{kipf2019compile}, RPL~\citep{gupta20relay}, OPAL~\citep{ajay2021opal}, LOVE~\citep{jiang2023efficient}, and PRISE~\citep{zheng2024prise}. 
They often operate in two stages: learning the primitives during the first and applying them to solve a downstream task during the second, possibly adapting the primitives in the process.
Compared with the visual state representation proposed in \ours, temporal action abstractions go in an orthogonal direction, and combining the benefits of both pretrained state representation and temporal action abstractions could be an exciting future direction.

\section{Additional Experiment Results}

\subsection{Finetuning}
\begin{figure}[!htbp]
\begin{centering}
\includegraphics[width=\textwidth]{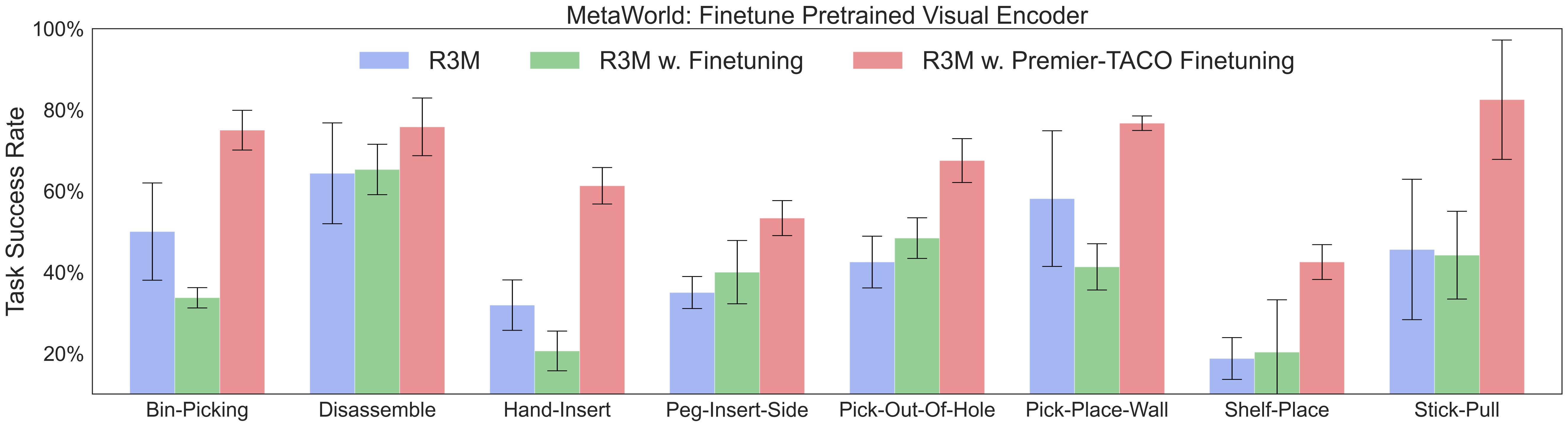} 
 \caption{\textbf{[(W4) Compatibility]} Finetune R3M~\citep{nair2022r3m}, a generalized Pretrained Visual Encoder with \ours learning objective vs. R3M with in-domain finetuning in \dmc and \mw.}\label{fig:r3m_metaworld}
 \vspace{-1.0em}
  \end{centering}
 \end{figure}

\begin{figure}[!htbp]
\begin{centering}
\includegraphics[width=\textwidth]{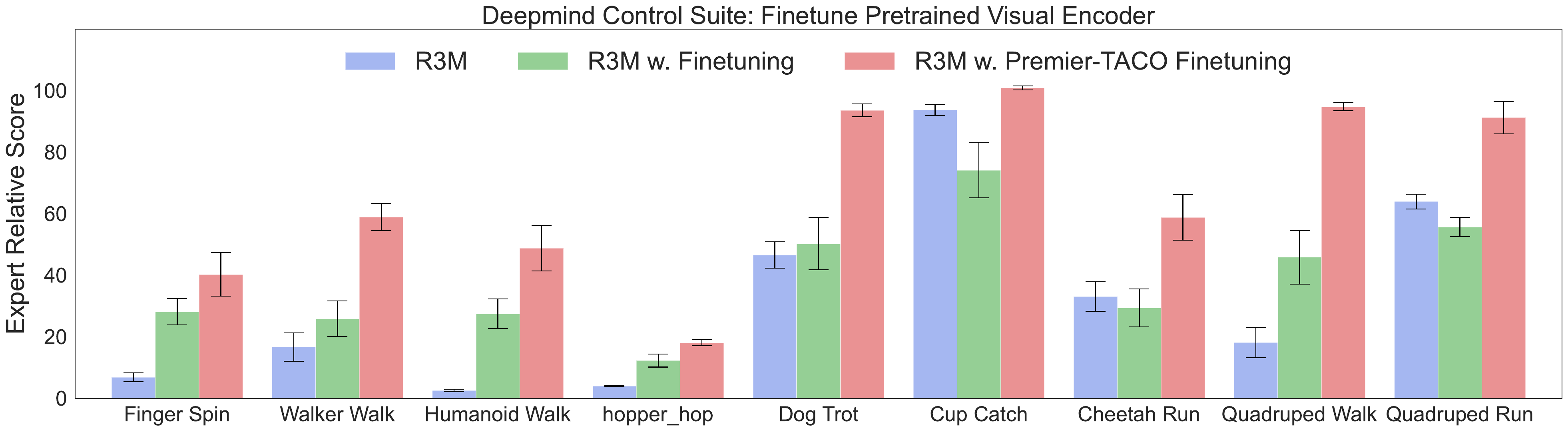} 
 \caption{\textbf{[(W4) Compatibility]} Finetune R3M~\citep{nair2022r3m}, a generalized Pretrained Visual Encoder with \ours learning objective vs. R3M with in-domain finetuning in \dmc and \mw.}\label{fig:r3m_dmc}
 \vspace{-1.0em}
  \end{centering}
 \end{figure}

Comparisons among R3M~\citep{nair2022r3m}, R3M with in-domain finetuning~\citep{hansen2022on} and R3M finetuned with \ours in \dmc and \mw are presented in Figure~\ref{fig:r3m_dmc} and \ref{fig:r3m_metaworld}.
\label{app:finetune}

\newpage
\subsection{Pretrained Visual Representations}
Here, we provide the full results for all pretrained visual encoders across all 18 tasks on Deepmind Control Suite and MetaWorld.
 \begin{table*}[htbp!]\small
\centering
\renewcommand{\arraystretch}{1.4}
\resizebox{\textwidth}{!}{%
\setlength{\tabcolsep}{3pt}
\begin{tabular}{ p{2.3cm}<{\centering} p{1.6cm}<{\centering} p{1.6cm}<{\centering} p{1.6cm}<{\centering} p{1.6cm}<{\centering}  p{2.3cm}<{\centering} p{1.6cm}<{\centering} p{1.6cm}<{\centering} p{1.6cm}<{\centering} p{1.6cm}<{\centering}}
\toprule
\multirow{2}{*}{\textbf{DMControl}} &   \multicolumn{4}{c}{Pretrained Visual Models} & \multirow{2}{*}{\textbf{\mw}} & \multicolumn{4}{c}{Pretrained Visual Models}\\  
 & PVR & MVP & R3M & VC-1 &   & PVR & MVP & R3M & VC-1 \\  
\hline
Finger Spin & $11.5 \pm 6.0 $ & $5.4 \pm 7.1 $ & $6.9 \pm 1.4$  &  $38.4 \pm 9.3$ & Bin Picking    &  $45.6\pm 5.6$ & $46.1 \pm 3.1$  & $50.0 \pm 12.0$  & $60.2 \pm 4.3$     \\
Hopper Hop &  $10.2\pm 1.5$  & $7.8 \pm 2.7 $&  $4.0 \pm 0.1$ & $23.2 \pm 4.9$   & Disassemble &  $47.6 \pm 5.8$   & $32.4\pm 5.1$  & $64.4 \pm 12.4$ &    $70.4 \pm 8.9 $     \\
Walker Walk &  $10.3 \pm 3.8$  &$ 8.30 \pm 1.6   $ &  $16.7 \pm 4.6$ & $  30.5\pm 6.2    $  & Hand Insert     &  $ 18.8 \pm 4.0   $ & $ 10.4  \pm  5.6   $ & $ 31.8 \pm 6.21  $  & $35.5 \pm 2.3 $     \\
Humanoid Walk  &  $ 7.6 \pm  3.4  $ &$ 3.2  \pm 0.5   $ &  $ 2.6 \pm 0.4   $ & $30.1 \pm 7.5$ & Peg Insert Side & $ 25.3 \pm 10.4   $  & $ 28.9 \pm 5.4   $ & $35.0\pm 3.95$ & $48.2 \pm 3.6  $ \\
Dog Trot        & $ 20.5 \pm 12.4 $ &  $ 32.9 \pm 6.0$       & $  46.6 \pm 4.3    $  &$73.5 \pm 6.4$ &          Pick Out Of Hole& $ 28.4 \pm 5.7    $  & $ 42.3  \pm 9.7   $ & $  42.5\pm 6.4    $  & $66.3 \pm 7.2$    \\
Cup Catch    & $ 60.2 \pm 10.3   $  & $ 56.7 \pm 8.9   $&  $93.7 \pm  1.8$  & $ 89.2 \pm 13.2   $ &    Pick Place Wall& $ 30.7 \pm 8.5   $  & $42.5  \pm 10.9   $ &   $ 58.1\pm 16.7 $     &$63.2 \pm 9.8 $    \\
Reacher Hard  &  $ 33.9 \pm 9.2   $ & $ 40.7 \pm 8.5   $ & $42.3\pm 5.6$ &  $64.9 \pm 5.8$  & Shelf Place  &  $ 19.5 \pm 6.4   $ & $ 21.2 \pm 8.3    $ &$  18.7\pm 5.15    $ &  $ 32.4 \pm 6.5$ \\
Cheetah Run     &  $ 26.7 \pm 3.8   $ & $27.3  \pm 4.4   $&  $  33.1 \pm 4.8    $ &    $39.5 \pm 9.7$       &    Stick Pull  & $30.2  \pm 4.6   $  &  $ 28.5 \pm 9.6   $&$  45.6\pm 17.3    $ & $52.4 \pm 5.6$ \\
Quadruped Walk    &  $  15.6 \pm 9.0   $ &$ 14.5 \pm 7.2    $ &  $  18.2 \pm 4.9    $ &   $63.2 \pm 4.0$     &  &   &  &   &  \\
Quadruped Run    &  $ 40.6 \pm 6.7    $ &$  43.2 \pm  4.2  $ & $  64.0 \pm 2.4    $  &     $61.3 \pm 8.5$    &  &   &  &   & \\
\bottomrule
\end{tabular}}
\caption{Few-shot results for pretrained visual representations~\citep{parisi22pvr, xiao2022mvp, nair2022r3m, majumdar2023vc1}}
\label{tab:pvrs}
\vspace{-1em}
\end{table*}

\subsection{LIBERO-10 success rate}
\begin{table*}[!h]\small
\centering
\renewcommand{\arraystretch}{1.4}
\resizebox{\textwidth}{!}{%
\setlength{\tabcolsep}{3pt}

\begin{tabular}{ p{2.3cm}<{\centering} p{1.7cm}<{\centering} p{1.6cm}<{\centering} p{1.6cm} <{\centering}  p{1.6cm}<{\centering} p{1.6cm}<{\centering} p{1.6cm}<{\centering} p{1.6cm}<{\centering} p{1.6cm}<{\centering} >{\columncolor{gray!25}}p{2.3cm}<{\centering}}
\toprule
\textbf{LIBERO}  &   \multicolumn{9}{c}{Models}   \\ 
   \textbf{Unseen Tasks} & LfS & Best PVRs & Multitask &  Inverse & CURL & ATC & SPR & TACO  & \textbf{\ours}\\  
\hline

Task 0 &  $19.9 \pm 4.1$ &  $ 21.2 \pm 3.5$   &  $23.3\pm 4.3$  &   $23.3\pm 4.3$ &    $16.7\pm 6.2$ &    $23.8\pm 6.9$ & $15.0\pm 0.0$ & $5.0\pm 0.0$ & $\bm{35.5\pm 7.5}$ \\
Task 1 &  $40.0 \pm 8.8$ &  $46.7 \pm 6.2$   &  $48.3\pm 10.3$  &   $38.3 \pm 9.2$ &    $26.7\pm 8.5$ &    $41.3\pm 7.5$ & $35.0\pm 6.3$ & $40.3\pm 4.1$ & $\bm{70.0\pm 5.0}$ \\
Task 2 &  $63.3 \pm 6.2$ &  $65.8 \pm 6.7$   &  $60.0\pm 4.1$  &   $51.6\pm 4.7$ &    $35.0\pm 4.1$ &    $65.0 \pm 9.3$ & $35.0\pm 5.0$ & $65.0\pm 9.3$ & $\bm{95.0\pm 7.2}$ \\
Task 3 &  $55.7 \pm 4.7$ &  $56.4 \pm 3.2$   &  $66.7\pm 8.4$  &   $70.1\pm 7.1$ &    $70.0\pm 6.8$ &    $\bm{83.8\pm 6.2}$ & $55.0\pm 5.4$ & $62.5\pm 7.3$ & $75.0\pm 13.2$ \\
Task 4 &  $\bm{43.3 \pm 6.2}$ &  $27.9 \pm 3.9$   &  $26.7\pm 3.1$  &   $28.3\pm 2.3$ &    $18.3\pm 2.5$ &    $25.0\pm 6.1$ & $23.7\pm 2.2$ & $15.5\pm 2.4$ & $30.7\pm 2.5$ \\
Task 5 &  $ 66.7 \pm 9.2$ &  $62.8 \pm 9.3$   &  $46.7\pm 3.8$  &   $63.3\pm 13.1$ &    $78.3\pm 2.4$ &    $78.8\pm 7.3$ & $68.7\pm 11.9$ & $52.3\pm 6.2$ & $\bm{80.0\pm 6.1}$ \\
Task 6 &  $6.7\pm 6.2$ &  $14.5 \pm 6.7$   &  $21.7\pm 2.3$  &   $11.6\pm 4.7$ &    $23.3\pm 2.4$ &    $11.2\pm 4.1$ & $12.5\pm 5.6$ & $19.8\pm 3.8$ & $\bm{27.5\pm 7.2}$ \\
Task 7 &  $26.7\pm 4.7$ &  $29.6 \pm 8.9$  &  $35.0\pm 7.1$  &   $38.3\pm 5.5$ &    $16.7\pm 2.4$ &    $26.3\pm 4.1$ & $35.0\pm 9.3$ & $22.3\pm 7.9$ & $\bm{50.3\pm 4.0}$ \\
\hline
\textbf{Mean} & $40.3$ &  $43.4$   &  $38.9$  &   $40.6$ &    $35.6$ &    $44.4$ & $35.0$ & $40.9$ & $\bm{58.0}$ \\
\bottomrule
\end{tabular}
}
\caption{\textbf{[(W1) Versatility (W2) Efficiency]} \textbf{Five-shot Behavior Cloning (BC) for unseen task of LIBERO.} Success rate of \ours and baselines across first 8 tasks on LIBERO-10. Results are aggregated over 4 random seeds. \textbf{Bold} numbers indicate the best results. }
\label{tab:libero_unseen_task}
\vspace{-1em}
\end{table*}
\newpage
\section{Additional Experiment Results on Downstream Online Reinforcement Learning}

\begin{figure}[h!]
\begin{subfigure}[h!]{0.48\textwidth}
\includegraphics[width=0.8\textwidth]{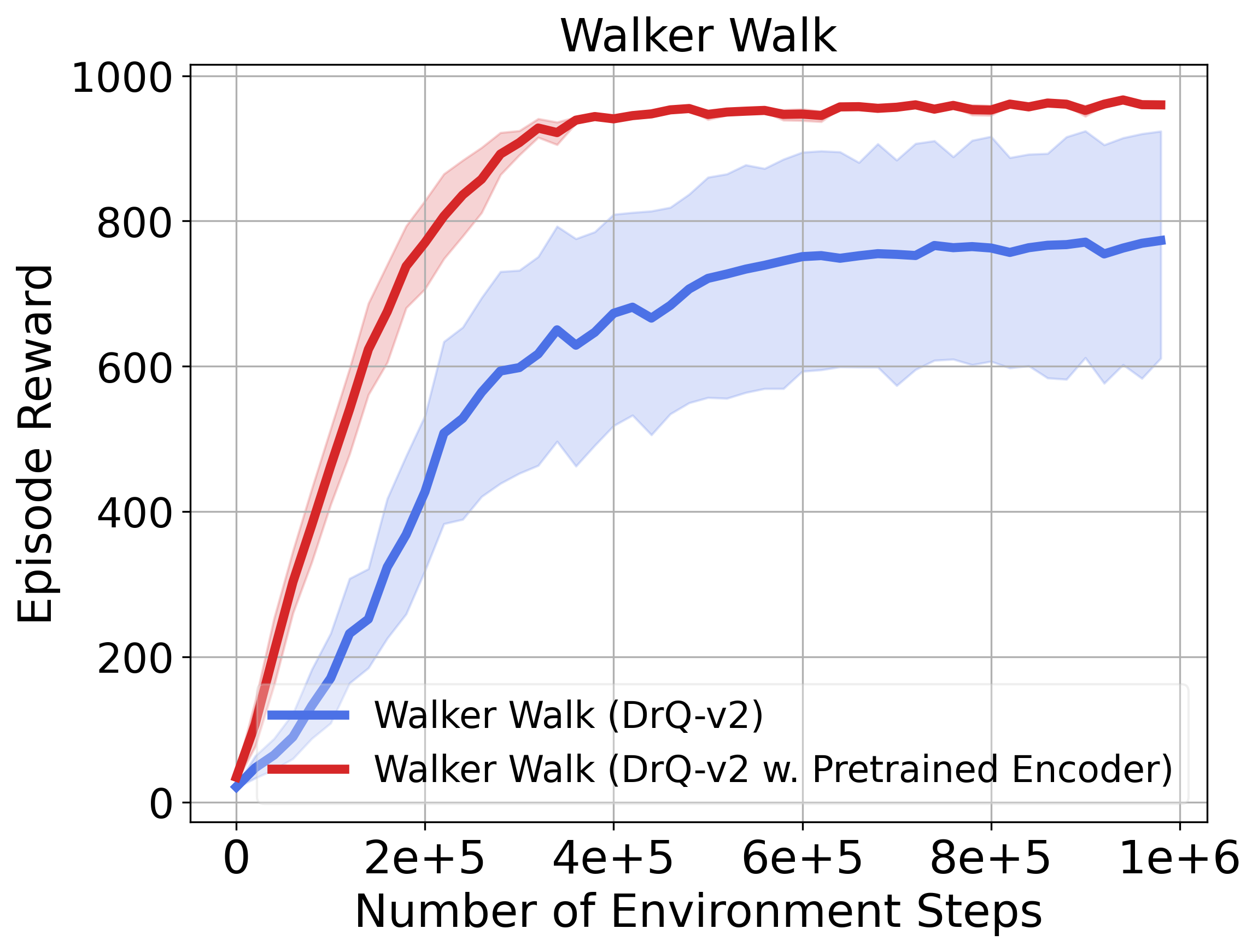}
\end{subfigure}
\begin{subfigure}[h!]{0.48\textwidth}
\includegraphics[width=0.8\textwidth]{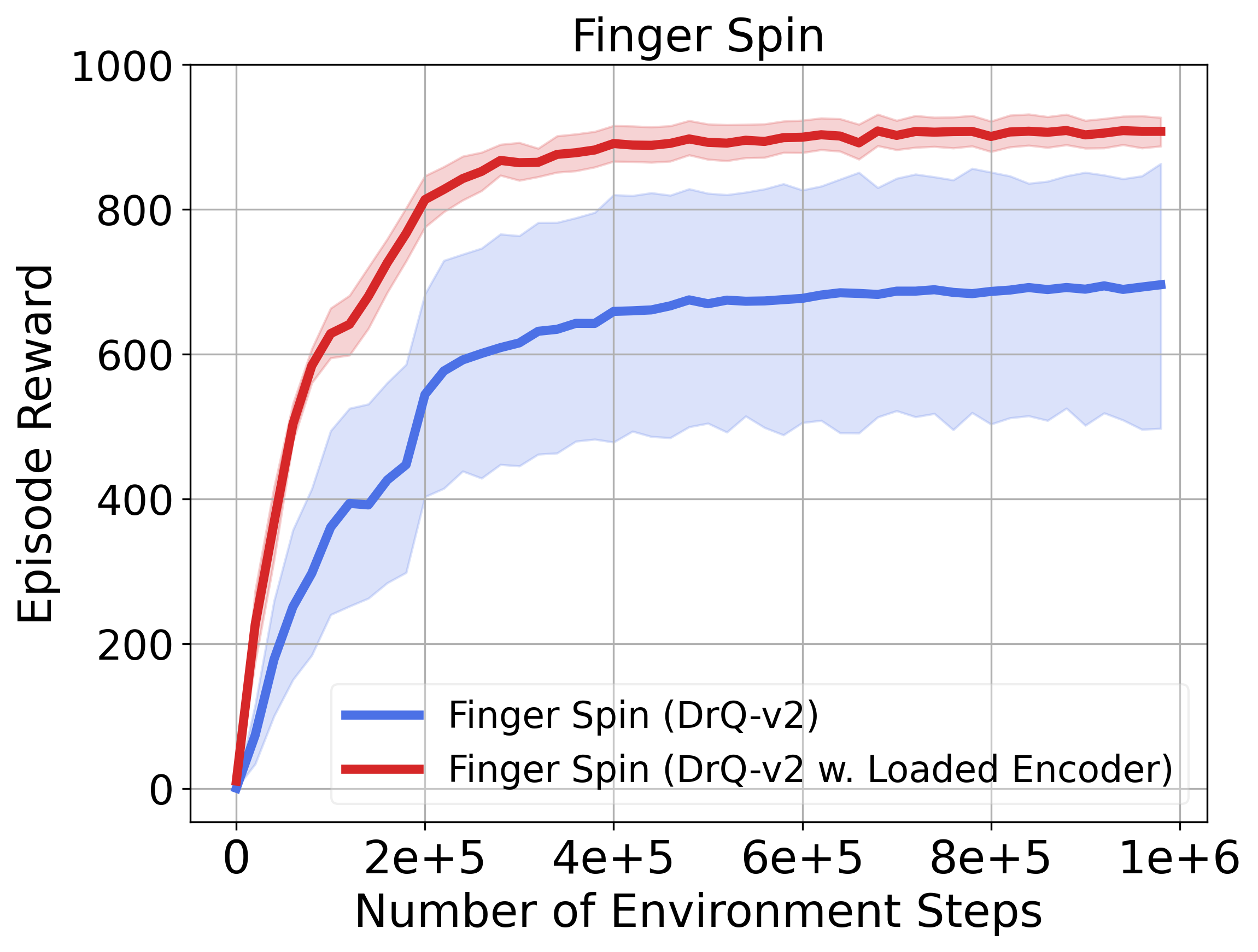}
\end{subfigure}
\caption{Downstream RL instead of imitation learning on Walker Walk (\textbf{Left}) and Finger Spin (\textbf{Right}). Results are aggregated over 8 random seeds.}
\label{fig:downstream_rl}
\end{figure}

While our paper primarily focuses on the sample-efficient imitation learning for downstream adaptation, we can also apply reinforcement learning instead of imitation learning. 
Here in Figure~\ref{fig:downstream_rl}, we have included additional experimental results on two unseen tasks, Walker Walk and Finger Spin, to showcase the downstream RL performance. 
We choose DrQ-v2~\cite{yarats2022drqv2} as the backbone visual RL algorithm and compare the performance of DrQ-v2 from scratch with DrQ-v2 using pretrained \ours encoder.
Notably, representation learned from pretrained \ours encoder can also significantly accelerate downstream RL learning. 
\section{Implementation Details}
\textbf{Dataset} For six pretraining tasks of Deepmind Control Suite, we train visual RL agents for individual tasks with DrQ-v2~\cite{yarats2022drqv2} until convergence, and we store all the historical interaction steps in a separate buffer. Then, we sample 200 trajectories from the buffer for all tasks except for Humanoid Stand and Dog Walk.
Since these two tasks are significantly harder, we use 1000 pretraining trajectories instead.  
Each episode in Deepmind Control Suite consists of 500 time steps.
In terms of the randomly collected dataset, we sample trajectories by taking actions with each dimension independently sampled from a uniform distribution $\mathcal{U}(-1.,1.)$.
For \mw, we collect 1000 trajectories for each task, where each episode consists of 200 time steps. 
We add a Gaussian noise of standard deviation 0.3 to the provided scripted policy.  
For LIBERO, we take the human demonstration dataset from~\citet{libero}, which contains 50 demosntration trajectories per task.

\textbf{Pretraining} For the shallow convolutional network used in Deepmind Control Suite and \mw, we follow the same architecture as in ~\citet{yarats2022drqv2} and add a layer normalization on top of the output of the ConvNet encoder. We set the feature dimension of the ConNet encoder to be 100. In total, this encoder has around 3.95 million parameters. 
\begin{lstlisting}[language=Python, caption=Shallow Convolutional Network Architecture Used in \ours]
class Encoder(nn.Module):
    def __init__(self):
        super().__init__()
        self.repr_dim = 32 * 35 * 35

        self.convnet = nn.Sequential(nn.Conv2d(84, 32, 3, stride=2),
                        nn.ReLU(), nn.Conv2d(32, 32, 3, stride=1),
                        nn.ReLU(), nn.Conv2d(32, 32, 3, stride=1),
                        nn.ReLU(), nn.Conv2d(32, 32, 3, stride=1),
                        nn.ReLU())
        self.trunk = nn.Sequential(nn.Linear(self.repr_dim, feature_dim),
                        nn.LayerNorm(feature_dim), nn.Tanh())

    def forward(self, obs):
        obs = obs / 255.0 - 0.5
        h = self.convnet(obs).view(h.shape[0], -1)
        return self.trunk(h)
\end{lstlisting}

For LIBERO, we use two randomly initialized (or pretrained) ResNet-18 encoders to encode the third-person view and first-person view images with FiLM~\citep{perez2018film} encoding method to incorporate the BERT embedding~\cite{bert} of the task language instruction. 
During downstream behavior cloning, we apply a transformer decoder module with context length 10 on top of the ResNet encodings to extract the temporal information, and then attach a two-layer MLP with hidden size 1024 as the policy head.
The architecture follows ResNet-T in~\citet{libero}.

For \ours loss, the number of timesteps $K$ is set to be 3 throughout the experiments, and the window size $W$ is fixed to be 5.
Action Encoder is a two-layer MLP with input size being the action space dimensionality, hidden size being 64, and output size being the same as the dimensionality of action space.
The projection layer $G$ is a two-layer MLP with input size being feature dimension plus the number of timesteps times the dimensionality of the action space. Its hidden size is 1024.
In terms of the projection layer $H$, it is also a two-layer MLP with input and output size both being the feature dimension and hidden size being 1024.
Throughout the experiments, we set the batch size to be 4096 and the learning rate to be 1e-4.
For the contrastive/self-supervised based baselines, CURL, ATC, and SPR, we use the same batch size of 4096 as \ours. 
For Multitask TD3+BC and Inverse dynamics modeling baselines, we use a batch size of 1024.

\textbf{Imitation Learning} A batch size of 128 and a learning rate of 1e-4 are used for Deepmind Control Suite and Metaworld, and a batch size of 64 is used for LIBERO. 
During behavior cloning, we finetune the Shallow ConvNet Encoder. 
However, when we applied \ours for the large pre-trained ResNet/ViT encoder models, we keep the model weights frozen.

In total, we take 100,000 gradient steps and conduct evaluations for every 1000 steps. For evaluations within the DeepMind Control Suite, we utilize the trained policy to execute 20 episodes, subsequently recording the mean episode reward. 
In the case of \mw and LIBERO, we execute 40 episodes and document the success rate of the trained policy.
We report the average of the highest three episode rewards/success rates from the 100 evaluated checkpoints.

\textbf{Computational Resources} For our experiments, we use 8 NVIDIA RTX A6000 with PyTorch Distributed DataParallel for pretraining visual representations, and we use NVIDIA RTX2080Ti for downstream imitation learning on Deepmind Control Suite and Metaworld, and RTX A5000 on LIBERO.

\newpage
\section{An Additional Ablation Study on Negative Example Sampling Strategy}

In \ours, we sample one negative example from a size $W$ window centered at the positive example for each data point. However, in principle, we could also use all samples within this window as negative examples instead of sampling only one.
In the table below, we compare the performance of two negative example sampling strategies across 10 unseen Deepmind Control Suite tasks. \textbf{Bold numbers} indicate the better results.

\begin{table}[h]
\centering
\begin{tabular}{l|c|c}
\hline
                & \textbf{Sampling 1} & \textbf{Sampling All} \\ \hline
Finger Spin    & \textbf{75.2 $\pm$ 0.6} & 70.2 $\pm$ 8.4\\ \hline
Hopper Hop     & $75.3 \pm 4.6$ & \textbf{76.1 $\pm$ 3.0} \\ \hline
Walker Walk    & {88.0 $\pm$ 0.8} & \textbf{88.5 $\pm$ 0.4} \\ \hline
Humanoid Walk  & {51.4 $\pm$ 4.9} & \textbf{56.4 $\pm$ 8.9} \\ \hline
Dog Trot       & \textbf{93.9 $\pm$ 5.4} & 92.1 $\pm$ 4.0 \\ \hline
Cup Catch      & \textbf{98.9 $\pm$ 0.1} & 98.3 $\pm$ 1.6 \\ \hline
Reacher Hard   & {\textbf{81.3 $\pm$ 1.8}} & 80.1 $\pm$ 5.8 \\ \hline
Cheetah Run    & {65.7 $\pm$ 1.1} & \textbf{69.3 $\pm$ 2.3} \\ \hline
Quadruped Walk & {83.2 $\pm$ 5.7} & \textbf{85.4 $\pm$ 4.2} \\ \hline
Quadruped Run  & {76.8 $\pm$ 7.5} & \textbf{82.1 $\pm$ 9.1} \\ \hline
Overall        & 79.0 & \textbf{79.8} \\ \hline
\end{tabular}
\caption{Results of two different negative sampling strategies on 10 unseen Deepmind Control Suite Tasks.}
\label{tab:negative_sampling}
\end{table}

As shown in Table~\ref{tab:negative_sampling}, we find that using all samples from the size $W$ window does not significantly enhance performance compared to \ours. Moreover, this approach considerably increases the computational overhead. Given these results, we chose a more computationally efficient strategy of sampling a single negative example from the size $W$ window in \ours.

\section{Task instructions of downstream LIBERO tasks}
Here in table~\ref{tab:libero_task_name}, we provide the language instruction for each of the LIBERO downstream task.
We refer readers to~\citep{libero} for more details of the tasks.
\vspace{-1.5em}
\begin{table}[!h]
\vskip 0.15in
\begin{center}
\begin{small}
\begin{sc}
\resizebox{1.0\columnwidth}{!}{%
\begin{tabular}{lcc}
\toprule
Task ID  & Task Scene & Task Instruction \\
\midrule
0 & living room scene2 & put both the alphabet soup and the tomato sauce in the basket \\
1 & living room scene2 & put both the cream cheese box and the butter in the basket \\
2 & kitchen scene3 & turn on the stove and put the moka pot on it \\
3 & kitchen scene4 & put the black bowl in the bottom drawer of the cabinet and close it \\
4 & living room scene5 & put the white mug on the left plate and put the yellow and white mug on the right plate \\
5 & study scene1 & pick up the book and place it in the back compartment of the caddy \\
6 & living room scene6 & put the white mug on the plate and put the chocolate pudding to the right of the plate \\
7 & living room scene1 & put both the alphabet soup and the cream cheese box in the basket \\
\bottomrule
\end{tabular}}
\caption{Language instructions for 8 LIBERO downstream tasks.}
\label{tab:libero_task_name}
\end{sc}
\end{small}
\end{center}
\vskip -0.1in
\end{table}

\end{document}